%% file: main.tex
\documentclass[10pt,twocolumn]{article}

\usepackage{iccv}
\usepackage{times}
\usepackage{epsfig}
\usepackage{graphicx}
\usepackage{amsmath}
\usepackage{amssymb}

\usepackage{multirow, subcaption}
\usepackage[linesnumbered,ruled,vlined]{algorithm2e}
\SetKwInput{KwInput}{Input}
\SetKwInput{KwOutput}{Output}
\usepackage[pagebackref=true,breaklinks=true,letterpaper=true,colorlinks,bookmarks=false]{hyperref}
\usepackage[pagebackref=true,breaklinks=true,colorlinks,bookmarks=false]{hyperref}
\usepackage{eso-pic}
\usepackage{booktabs}

\usepackage[font=normalsize]{caption}
\usepackage{commath}

\usepackage[breaklinks=true,bookmarks=false]{hyperref}

\iccvfinalcopy 

\def\iccvPaperID{****} 

\begin{document}

\title{Camera Distortion-aware 3D Human Pose Estimation in Video\\with Optimization-based Meta-Learning}

\author{Hanbyel Cho ~~~~~~~~~~~~~~~ Yooshin Cho ~~~~~~~~~~~~~~~ Jaemyung Yu ~~~~~~~~~~~~~~~ Junmo Kim\\
        School of Electrical Engineering, KAIST, South Korea \\
        {\tt\small \{tlrl4658, choys95, jaemyung, junmo.kim\}@kaist.ac.kr}
        }

\maketitle

\input{body}

{\small
\bibliographystyle{ieee_fullname}
\bibliography{egbib}
}

\input{supplementary.tex}

\end{document}

%% file: body.tex
\begin{abstract}
    Existing 3D human pose estimation algorithms trained on distortion-free datasets suffer performance drop when applied to new scenarios with a specific camera distortion. In this paper, we propose a simple yet effective model for 3D human pose estimation in video that can quickly adapt to any distortion environment by utilizing MAML, a representative optimization-based meta-learning algorithm. We consider a sequence of 2D keypoints in a particular distortion as a single task of MAML. However, due to the absence of a large-scale dataset in a distorted environment, we propose an efficient method to generate synthetic distorted data from undistorted 2D keypoints. For the evaluation, we assume two practical testing situations depending on whether a motion capture sensor is available or not. In particular, we propose Inference Stage Optimization using bone-length symmetry and consistency.
    Extensive evaluation shows that our proposed method successfully adapts to various degrees of distortion in the testing phase and outperforms the existing state-of-the-art approaches. The proposed method is useful in practice because it does not require camera calibration and additional computations in a testing set-up. Code is available at \url{https://github.com/hanbyel0105/CamDistHumanPose3D}.
\end{abstract}
\vspace{-2mm}

\section{Introduction}
    3D human pose estimation is a task that localizes 3D human body joint from an RGB input. As a fundamental task in computer vision, it is applied to many downstream applications, \eg, action recognition~\cite{ref1_SGN_CVPR2020,ref2_global-context-aware_CVPR2017,ref3_action_CVPR2017}, human body reconstruction~\cite{ref4_8565937,ref5_ijcai2018-105}, and human-computer interaction~\cite{ref6_10.5555/3027754}. Particularly, 3D pose estimation for monocular video, which predicts 3D joint in inputs from a single-camera, has attracted a lot of academic interest recently~\cite{ref7_martinez2017simple,ref8_zhao2019semantic,ref9_pavllo20193d,ref10_cai2019exploiting, ref11_anatomy3D, ref12_liu2020attention} because of the simplicity of the hardware setting in use and its advantage of being able to leverage temporal information to resolve inherent depth ambiguity.

    \begin{figure}[t]
    	\begin{center}
    		\captionsetup{justification=centering}
    		\begin{subfigure}[t]{0.325\linewidth}
    			\centering
    			\includegraphics[width=1\columnwidth]{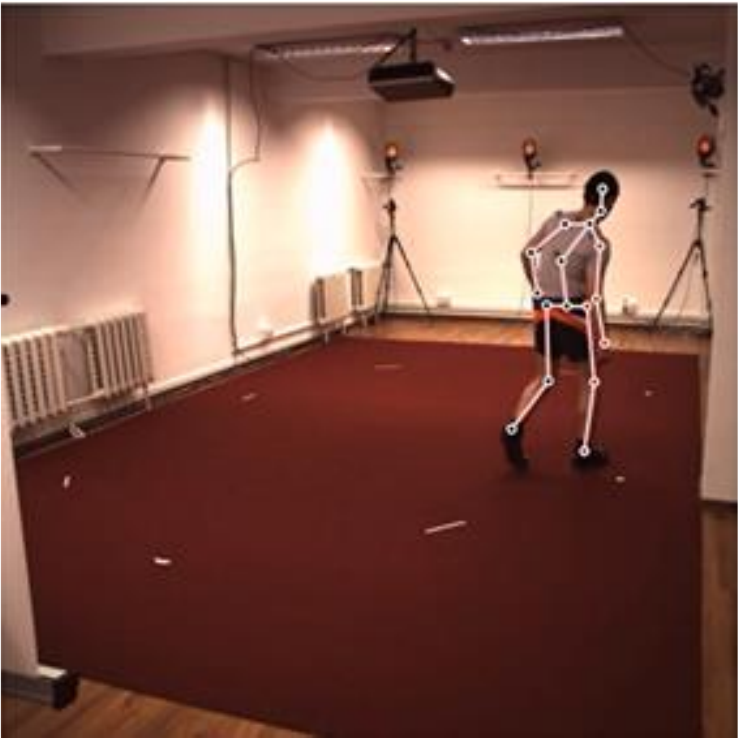}
    		\end{subfigure}
    		\begin{subfigure}[t]{0.325\linewidth}
    			\centering
    			\includegraphics[width=1\columnwidth]{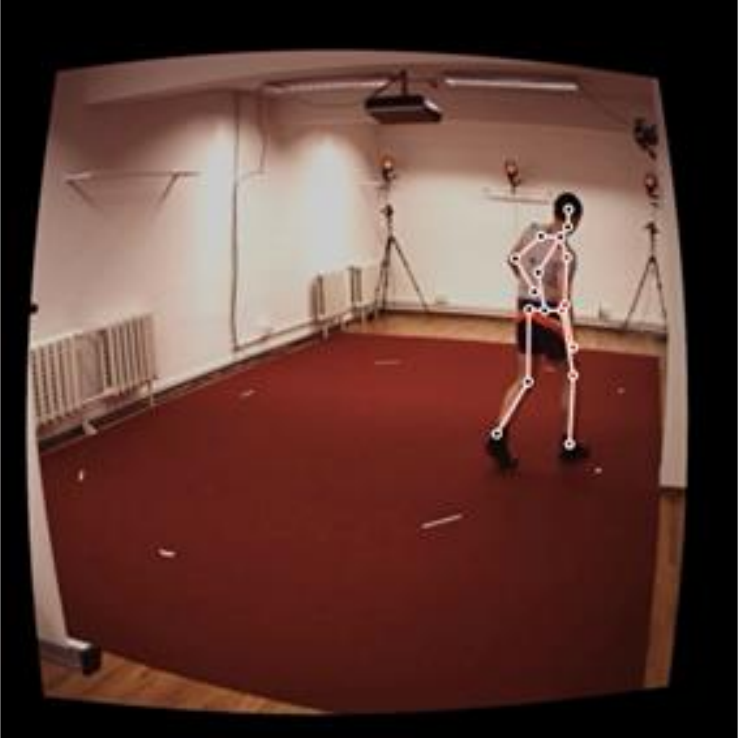}
    		\end{subfigure}
    		\begin{subfigure}[t]{0.325\linewidth}
    			\centering
    			\includegraphics[width=1\columnwidth]{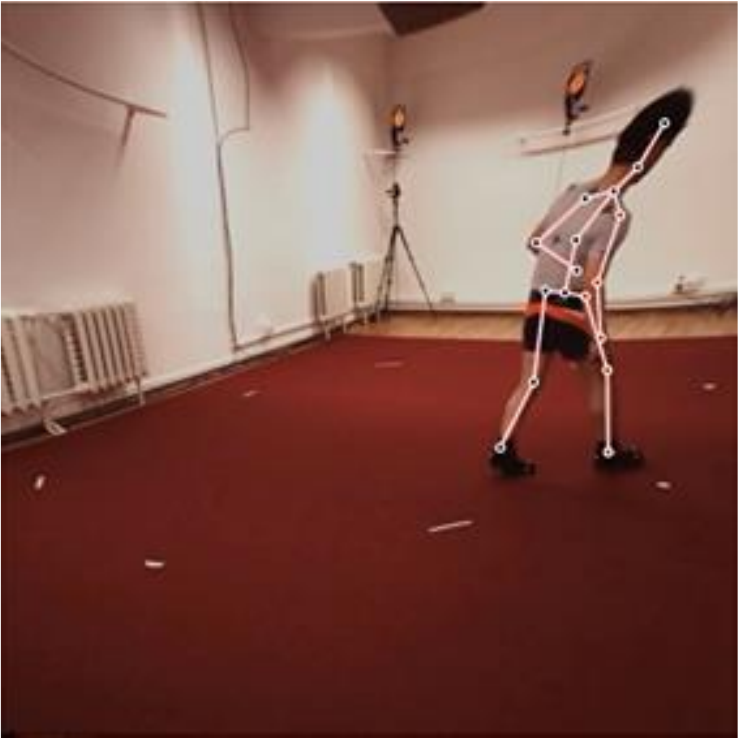}
    		\end{subfigure}
    		
    		\begin{subfigure}[t]{0.325\linewidth}
    			\centering
    			\includegraphics[width=1\columnwidth]{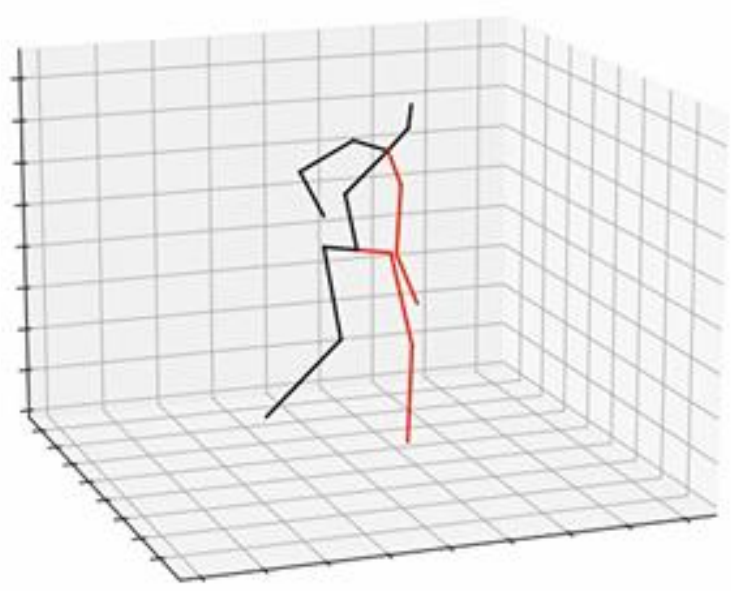}
    			\caption{Undistorted}
    		\end{subfigure}
    		\begin{subfigure}[t]{0.325\linewidth}
    			\centering
    			\includegraphics[width=1\columnwidth]{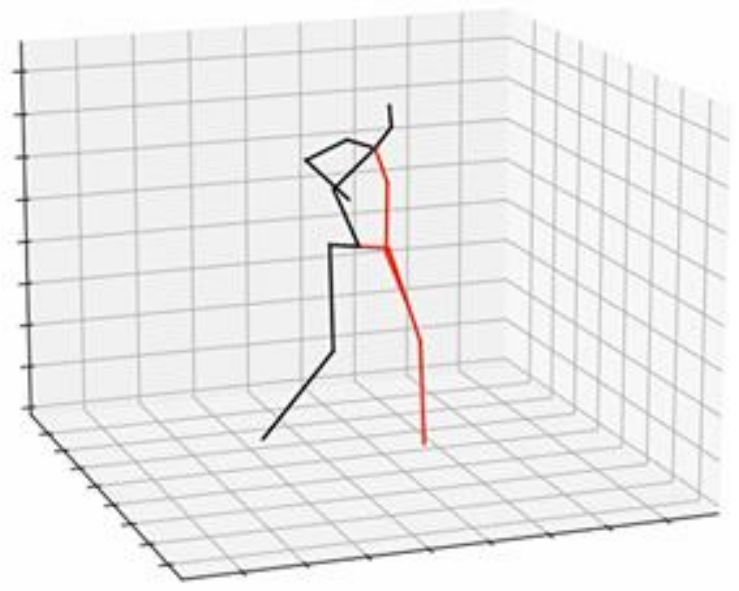}
    			\caption{Distortion 1}
    		\end{subfigure}
    		\begin{subfigure}[t]{0.325\linewidth}
    			\centering
    			\includegraphics[width=1\columnwidth]{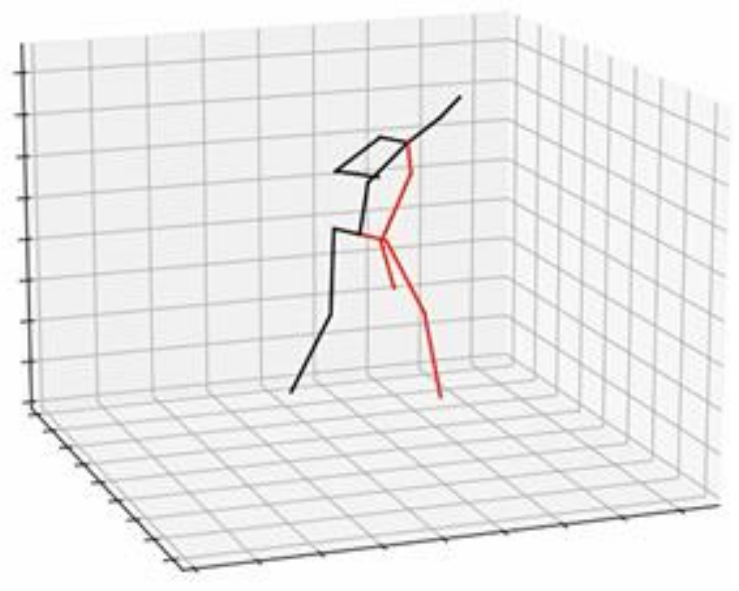}
    			\caption{Distortion 2}
    		\end{subfigure}
    	\end{center}
    	    \vspace{-3mm}
    		\caption{3D reconstruction for videos with varying degrees of distortion using a network trained with a distortion-free dataset. \textbf{Top:} input video frames with 2D pose overlay. \textbf{Bottom:} 3D reconstruction. 3D reconstruction of (a) is predicted from undistorted video, and (b) and (c) are predicted from video with different degrees of distortion, respectively.}
    	\label{fig:performance_drop}
    	\vspace{-1mm}
    \end{figure}
    
    \begin{table}[t]
    	\centering
    	\small
    	\tabcolsep=1mm
    	\resizebox{\columnwidth}{!}{
    		\begin{tabular}{@{}l|r|r|r@{}}
    		Condition & ~~~~~~~~~~~MPJPE($\downarrow$) & ~~~~~~~~~~~P-MPJPE($\downarrow$) & ~~~~~~~~~~~PCKh@0.5($\uparrow$) \\
    		\midrule
            Undistorted     &	48.5             & 37.1           & 87.1\\
            Distortion 1    &   94.4(+45.9)      & 65.6(+28.5)    & 57.7(-29.4)\\
            Distortion 2    &   133.8(+85.3)     & 79.2(+42.1)    & 38.2(-48.9)\\        
    		\bottomrule
    		\end{tabular}
    		}
    	\caption{Performance drop in environments with distortion of a network trained with a distortion-free dataset. \small\emph{Distortion 1} \normalsize and \small\emph{Distortion 2} \normalsize  are the cases of barrel distortion and pincushion distortion, with tangential distortion, respectively.}
    	\label{tbl:performance_drop}
    \end{table}

    Recently, many state-of-the-art studies adopted two-stage architecture to achieve higher performance. In this architecture, 2D keypoints are first extracted from the off-the-shelf 2D keypoint detector~\cite{ref13_Chen_2018_CVPR,ref14_He_2017_ICCV,ref15_870a37a4f4e24a66bce3afcddaadf871,ref16_Sun_2019_CVPR}, and 3D keypoints are inferred using the predicted 2D keypoints sequence as input. These approaches simplify the 3D pose estimation problem to solve depth ambiguity from 2D joint sequences. This allows the study~\cite{ref17_DeepKinematics_CVPR2020,ref11_anatomy3D,ref8_zhao2019semantic} of algorithms explicitly using information such as skeleton kinematics and motion of human, which showed plausible results.

    Despite significant advances in 2D-keypoint-based 3D pose estimation, there still remain certain limitations. That is existing 3D human pose estimation algorithms trained on distortion-free datasets show severe performance drop when applied to new scenarios with a specific camera distortion, as shown in Figure~\ref{fig:performance_drop} and Table~\ref{tbl:performance_drop}. Previously, preprocessed images were used when inferring 3D joints from distorted inputs. However, it is important to make models that can adapt themselves to arbitrary distortion in the testing phase, as algorithms that are needed in preprocessing, such as camera calibration, are sometimes difficult to apply and they also introduce certain errors of their own. This is substantially important issue in the wild use of the algorithms; however, cross-scenario research on camera distortion has been out of scope due to the absence of a dataset with various degrees of camera distortion.

    To overcome this limitation, in this work, we propose a simple yet effective model for 3D human pose estimation in video that can quickly adapt to any distortion environment by utilizing model-agnostic meta-learning (MAML)~\cite{ref20_pmlr-v70-finn17a}, a representative optimization-based meta-learning algorithm. We focus on training a distorted-2D-keypoints-conditioned 3D pose estimator to be able to quickly adapt to camera distortion, because we found that 2D keypoint detector is good at finding distorted 2D keypoints consistent with distorted images. Therefore, we consider a sequence of 2D keypoints in a particular distortion as a single task of MAML. However, due to the absence of a large-scale dataset with a distorted environment, we propose an efficient method to generate synthetic distorted data from undistorted 2D keypoints. Note that, the goal of training phase is not to just increase the performance at a particular distortion, but to train a network sensitive to distortion, allowing the network to adapt quickly to arbitrary distortion in the testing phase. For the testing phase, the trained network is first adapted to a specific camera distortion environment by fine-tuning or Inference Stage Optimization, which as proposed in recent work~\cite{ref18_ISO_NeurIPS2020} in the following two scenarios.

    For evaluation, we assume two practical situations in which the proposed method will be used and confirm that our algorithm is useful for each case. First, \textit{Scenario 1} is a situation in which a user can collect data using motion capture sensors in front of a testing environment, as shown in Figure~\ref{fig:overall} (b). In this case, data with the same distortion as the testing environment can be obtained, but it would be in relatively small amounts compared with a large-scale dataset (\eg, Human3.6M~\cite{ref19_6682899}) collected in the laboratory environment. Therefore, it is important to transfer knowledge trained with a large-scale dataset as much as possible. To validate the usefulness of the proposed method, we construct a small-scale dataset with the same distortion as the test environment, and evaluate whether the network can adapt well through naive fine-tuning.

    Second, \textit{Scenario 2} is when the user is unable to obtain data in a testing environment, as shown in Figure~\ref{fig:overall} (c). In this case, the network should be adapted to specific distortions using only test videos. In a recent study~\cite{ref18_ISO_NeurIPS2020}, the authors proposed the concept named Inference Stage Optimization (ISO) to adapt network using only test data before testing. We also utilize ISO in this case. To this end, we propose a novel ISO method based on skeleton symmetry and consistency. This might be a weak constraint, but we confirm that our network is fully adaptable even with these constraints because it has been sensitively trained on distortion.

    In summary, our overall contribution is four-fold:
    \vspace{-1mm}
    \begin{itemize}
        \item To the best of our knowledge, our method, which utilized optimization-based meta-learning, is the first algorithm that can adapt to arbitrary camera distortion at the testing phase.
        \vspace{-2mm}
        \item We propose an efficient method to generate synthetic distorted data from undistorted 2D keypoints, enabling cross-scenario research on camera distortion, which has been out-of-scope due to the absence of datasets with distortion.
        \vspace{-2mm}
        \item We validate the effectiveness of the proposed method for each case, assuming two practical testing environments. In particular, we propose the ISO method using bone-length symmetry and consistency.
        \vspace{-2mm}
        \item Our proposed method is useful in practical applications because it does not require calibration for the testing camera and additional computational complexity.
    \end{itemize}

\section{Related Work}
    \subsection{3D Human Pose Estimation}
        \vspace{-1mm}
        Since the success of 2D human pose estimation, 3D human pose estimation has been widely studied. Martinez~\etal~\cite{ref7_martinez2017simple} successfully predicted 3D poses from 2D joint locations using simple and lightweight networks. It showed better results than previous studies that involved training with raw image pixels. To make better use of 2D keypoints, GCN and attention mechanism were applied to learn the global relationship between joints \cite{ref8_zhao2019semantic,ref12_liu2020attention}. In contrast, Pavllo~\etal~\cite{ref9_pavllo20193d} predicted 3D poses using video to overcome inherent ambiguity that multiple 3D poses can be mapped to the same 2D pose. Furthermore, prior knowledge about human body structure was explicitly utilized to give constraints ~\cite{ref10_cai2019exploiting,ref11_anatomy3D}. Despite substantial progress in this field, the performance severely drops when camera distortion occurs due to the changes in camera parameters in test environment.
    
    \subsection{Cross-scenario Pose Estimation}
        \vspace{-1mm}
        Deep learning models have substantially improved in recent years. However, due to the limitation of supervised learning on datasets that lack diversity, even state-of-the-art algorithms show poor results in-the-wild. To be robust on a domain gap between training and inference, many studies have been conducted~\cite{ref21_shocher2018zero,ref22_soh2020meta,ref18_ISO_NeurIPS2020}. A recent study~\cite{ref18_ISO_NeurIPS2020} proposed the domain (\eg, varying poses, camera viewpoints, body size, and appearances) robust 3D pose estimation algorithm that adapts to the target domain using self-supervised learning schemes named Inference Stage Optimization (ISO) using cycle consistency among 2D and 3D spaces. In this paper, we focus on the domain gap of camera distortion caused by the different camera settings at the testing phase, which has been out-of-scope.
    
    \subsection{Optimization-based Meta-Learning}
        \vspace{-1mm}
        There are three common categories in meta-learning. The first category is the metric-based approach~\cite{ref23_Koch2015SiameseNN,ref24_NIPS2016_90e13578,ref25_NIPS2017_cb8da676,ref26_Sung_2018_CVPR}, which learns a good metric that expresses the relationship between inputs in task space and applies it well to new samples. The second category is a model-based approach~\cite{ref27_pmlr-v48-santoro16,ref28_pmlr-v70-munkhdalai17a,ref29_mishra2018a} that controls the structure of a target model through another model called meta-learner. The last category is an optimization-based approach~\cite{ref30_10.5555/3157382.3157543,ref31_Ravi2017OptimizationAA,ref20_pmlr-v70-finn17a,ref32_NEURIPS2019_072b030b} that looks for sensitive initial parameters for tasks and quickly adapts to new tasks with only a few samples. In this work, we utilize MAML~\cite{ref20_pmlr-v70-finn17a}, which belongs to the optimization-based approach so that the network can adapt quickly to arbitrary camera distortion in the testing phase.

\section{Preliminary}
    In this section, we introduce background knowledge on camera distortion and framework of the MAML algorithm.
    \vspace{-7mm}

\begin{figure}[t]
        	\begin{center}
        		\captionsetup{justification=centering}
        		\begin{subfigure}[t]{0.23\linewidth}
        			\centering
        			\includegraphics[width=1\columnwidth]{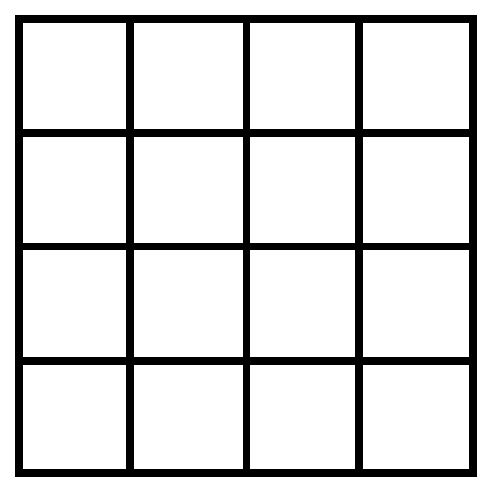}
        			\vspace{-6mm}
        			\caption{\footnotesize Undistorted}
        		\end{subfigure}
        		\begin{subfigure}[t]{0.23\linewidth}
        			\centering
        			\includegraphics[width=1\columnwidth]{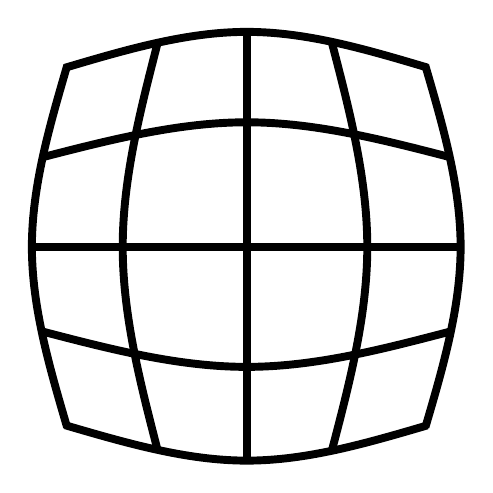}
        			\vspace{-6mm}
        			\caption{\footnotesize Barrel}
        		\end{subfigure}
        		\begin{subfigure}[t]{0.23\linewidth}
        			\centering
        			\includegraphics[width=1\columnwidth]{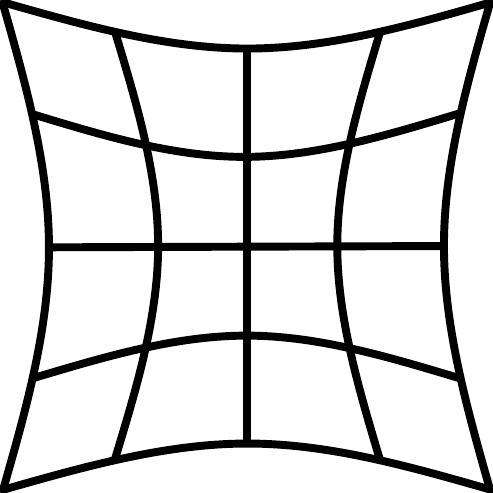}
        			\vspace{-6mm}
        			\caption{\footnotesize Pincushion}
        		\end{subfigure}
    
        		\begin{subfigure}[t]{0.23\linewidth}
        			\centering
        			\includegraphics[width=1\columnwidth]{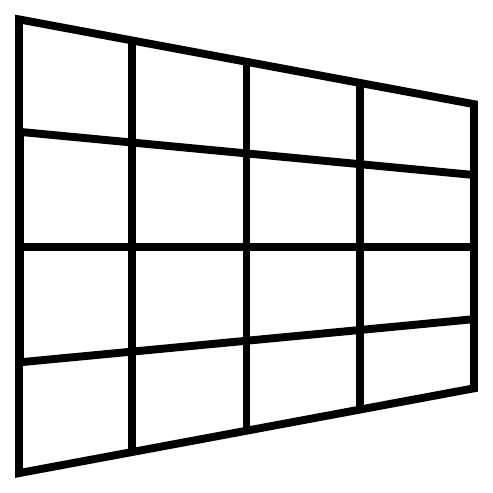}
        			\vspace{-6mm}
        			\caption{\footnotesize Tangential $x$}
        		\end{subfigure}
        		\begin{subfigure}[t]{0.23\linewidth}
        			\centering
        			\includegraphics[width=1\columnwidth]{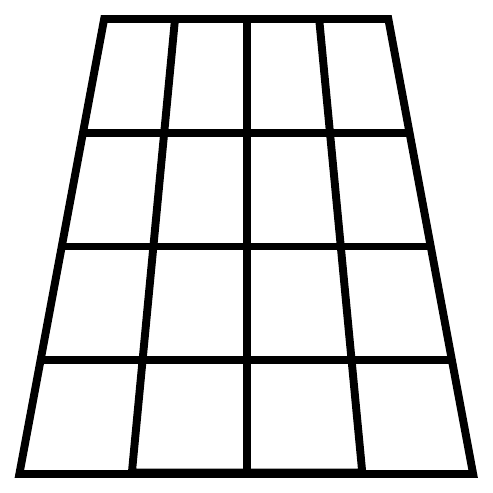}
        			\vspace{-6mm}
        			\caption{\footnotesize Tangential $y$}
        		\end{subfigure}
        	\end{center}
        	    \vspace{-4mm}
        		\caption{Types of camera distortion. (b) and (c) represent radial distortion, and (d) and (e) represent tangential distortion. Radial distortion and tangential distortion can occur simultaneously.}
        		\vspace{-4mm}
        	\label{fig:camera distortion}
        \end{figure}

    \paragraph{Camera Distortion.}
        There are two kinds of camera distortion. The first is radial distortion, which is caused by the refractive index of the convex lens. Radial distortion is determined by the distance from the center of the image, and it is usually expressed in parameters $k_1$, $k_2$, and $k_3$. The value of $k_1$ determines the main form of the distortion. A negative $k_1$ and a positive $k_1$ result in barrel distortion and pincushion distortion respectively, as shown in Figure~\ref{fig:camera distortion} (b) and (c). The second is tangential distortion, which is caused by the misalignment of the camera lens and the image sensor (\eg, CCD and CMOS) during manufacturing of the camera. This can be approximated by parameters $p_1$ and $p_2$. The $p_1$ and $p_2$ cause ladder shape distortion mainly in $x$-axis and $y$-axis respectively, as shown in Figure~\ref{fig:camera distortion} (d) and (e). Both distortions are common in commercial cameras, and radial distortion is particularly severe in wide-angle cameras.

        \vspace{-2mm}

    \paragraph{Model-Agnostic Meta-Learning.}
        The stage of meta-learning consists of meta-training and meta-testing. We consider a model represented by a function $g_\theta$ with parameters $\theta$, that outputs $\mathbf{y}$ with input as $\mathbf{x}$. The objective of meta-training is to find initial transferable weights that can be adapt to new tasks. For meta-training, a batch of tasks $\mathcal{T}_i$ is sampled from task distribution $p(\mathcal{T})$. The model is first optimized through the task-specific loss $\mathcal{L}_{\mathcal{T}_{i}}$ using training samples within a task (\textit{task-level training}), and meta-optimization is performed using test samples (\textit{task-level testing}). In meta-testing, the model adapts to a new task $\mathcal{T}_{new}$ using only a few samples. In this study, we use MAML~\cite{ref20_pmlr-v70-finn17a}, in which input $\mathbf{x}$ and output $\mathbf{y}$ are distorted 2D keypoint trajectory and 3D joints respectively. Various distortion parameters construct task distribution, and each task corresponds to a 3D pose estimation from the distorted 2D keypoint trajectory with a particular distortion parameter.
\vspace{-1mm}

\section{Method}
\begin{figure*}
        \begin{center}
            \includegraphics[width=1\linewidth]{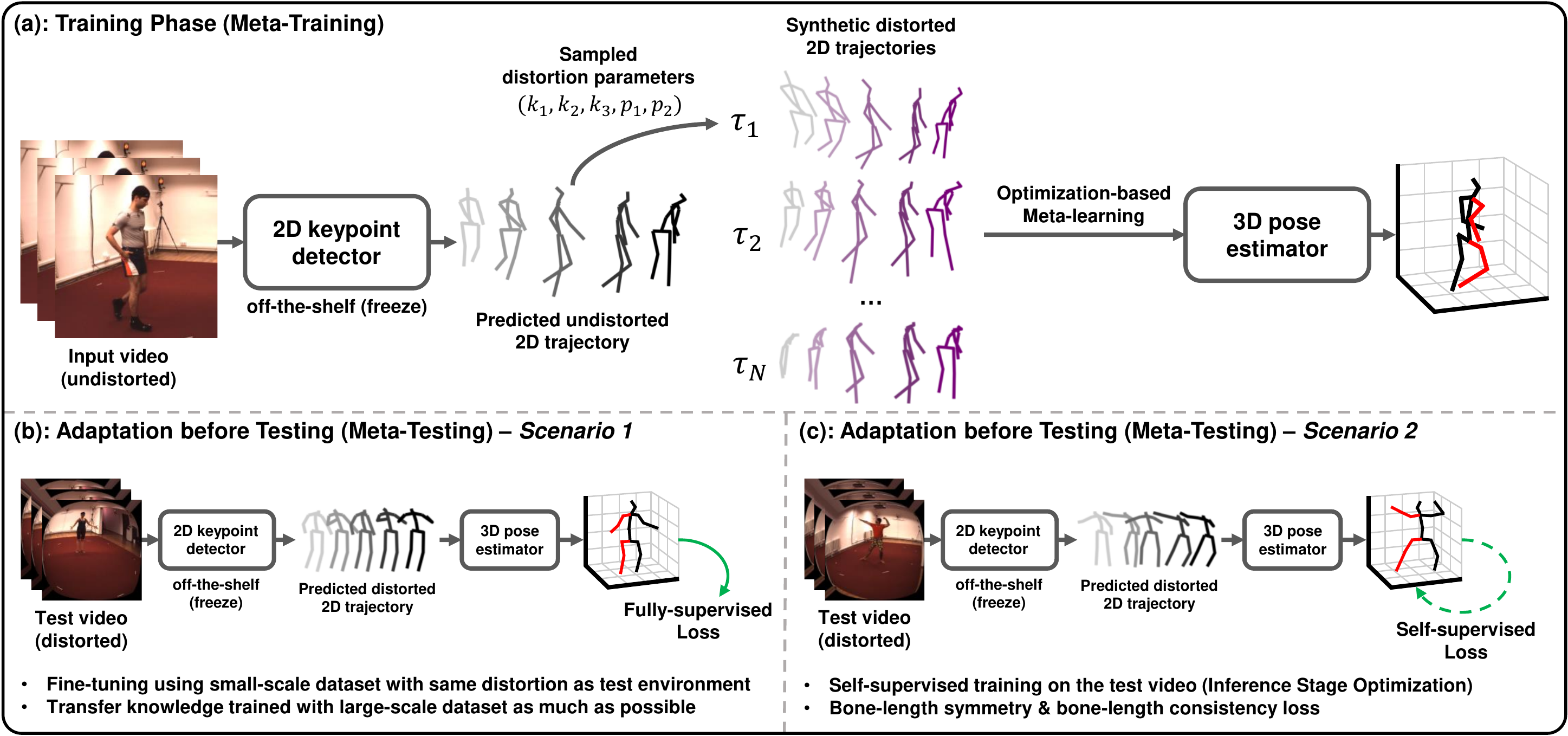}
        \end{center}
        \vspace{-4.5mm}
        \caption{Overall framework of our methods. (a) We train a 2D-keypoint-conditioned 3D pose estimator that can quickly adapt to any distortions using only an undistorted large-scale dataset. Before the trained network can be used in practice, it must be adapted to a certain distortion. (b) and (c) represent adaptation method for \small$\textit{Scenario 1}$ \normalsize and \small$\textit{Scenario 2}$\normalsize, respectively.}
        \vspace{-1mm}
        \label{fig:overall}
        \vspace{-2.5mm}
    \end{figure*}

    \vspace{-1mm}
    The overall framework of our method is shown in Figure~\ref{fig:overall}. In this section, we first propose a method for \textit{generating synthetic distorted tasks}. Then, we describe two phases that constitute our method: \textit{training phase} and \textit{adaptation before testing}.
    \vspace{-1mm}
    
    \subsection{Synthetic Distorted Task Generation}
        \label{sec1}
        We found that 2D keypoint detector is good at finding distorted 2D keypoints consistent with distorted images, as shown in the top row of Figure~\ref{fig:performance_drop} because it is based on the texture of images. Thus, in the training phase, our goal is to train a 3D pose estimator conditioned on distorted 2D detection to be able to quickly adapt to various distortions by applying meta-learning. Meta-learning in our case requires tasks under varying degrees of distortion. In this section, we describe how to efficiently generate distorted tasks from undistorted videos.
        
        Given a video clip with frame length of $T$, first 2D keypoints are obtained by a pretrained 2D keypoint detector (\eg, Mask R-CNN~\cite{ref14_He_2017_ICCV}). Let $\tilde{\mathbf{p}}_t\in\mathbb{R}^{J\times2}$ denotes predicted 2D coordinates of $J$ keypoints of the human in the frame and $\tilde{\mathbf{P}} = \{\tilde{\mathbf{p}}_t\}_{t=1}^T$ denotes the set of joints for a video clip. Specifically, $\tilde{\mathbf{p}}_t = \{[\tilde{a}_{t,j},\tilde{b}_{t,j}]\}_{j=1}^J$ where $\tilde{a}_{t,j}$ and $\tilde{b}_{t,j}$ denote $x$ and $y$ coordinates of $j$th joint at frame $t$, respectively.
        
\begin{figure}[t]
        	\begin{center}
        		\captionsetup{justification=centering}
        		\begin{subfigure}[t]{1.0\linewidth}
        			\centering
        			\includegraphics[width=1\columnwidth]{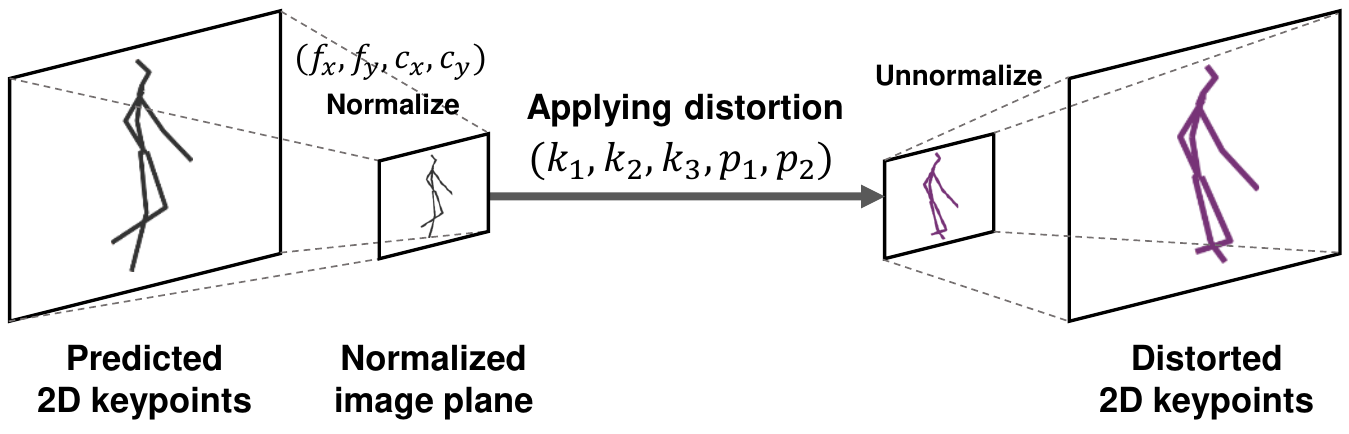}
        			\caption{Generating distorted 2D keypoints from predicted ones.}
        		\end{subfigure}
        		\begin{subfigure}[t]{1\linewidth}
        			\centering
        			\includegraphics[width=1\columnwidth]{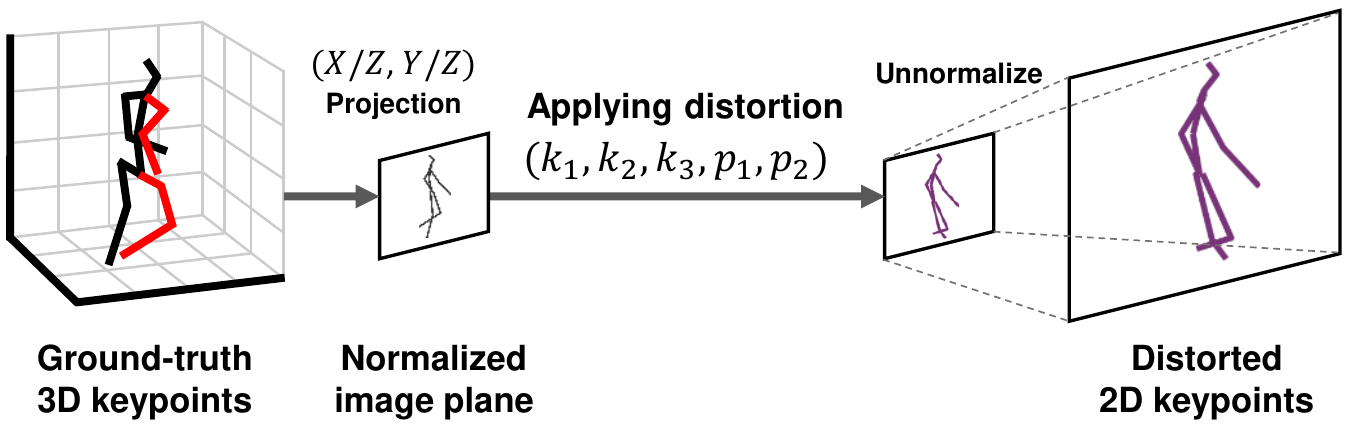}
        			\caption{Generating distorted 2D keypoints from 3D ground-truth.}
        			\label{subfig:b}
        		\end{subfigure}
        	\end{center}
        	    \vspace{-5mm}
        		\caption{Methods to generate distorted 2D keypoints.}
        	    \vspace{-4mm}
        	\label{fig:synthetic data}
        \end{figure}

        To generate synthetic distorted tasks, we apply the camera distortion model~\cite{ref38_CamCalibration} directly to predicted 2D keypoints. We omit subscript $t$ and $j$ for simplicity. As shown in Figure~\ref{fig:synthetic data} (a), the process of generating the task with particular distortion parameters (\ie, $k_1,k_2,k_3,p_1,p_2$) is divided into three steps. The first is obtaining normalized 2D keypoints (denoted as $[\tilde{a}_n,\tilde{b}_n]$) and distance between the point and image center (denoted as $r$). As camera distortion models should be applied on a normalized image plane, we first normalize 2D keypoints with focal length (denoted as $\mathbf{f}=[f_x,f_y]$) and optical center (denoted as $\mathbf{c}=[c_x,c_y]$) using the following equations:
        \begin{equation}
            \small
            \tilde{a}_n = \frac{\tilde{a}-c_x}{f_x},\tilde{b}_n = \frac{\tilde{b}-c_y}{f_y},r=\sqrt{\tilde{a}_n^2+\tilde{b}_n^2}.
            \label{eq:hb_1}
        \end{equation}
        Then, we apply distortion to the normalized 2D keypoints using the following equations:
        \begin{equation}
            \small
            \tilde{a}_{n,d} = \tilde{a}_n(d_r+d_t)+p_1r^2,\tilde{b}_{n,d} = \tilde{b}_n(d_r+d_t)+p_2r^2,
            \label{eq:hb_3}
        \end{equation}
        where intermediate variable $d_r$ and $d_t$ are obtained by \small$d_r = 1+k_1r^2+k_2r^4+k_3r^6$ \normalsize and \small$d_t = 2p_1\tilde{a}_n+2p_2\tilde{b}_n$ \normalsize respectively. Finally, distorted 2D keypoints (denoted as $[\tilde{a}_d,\tilde{b}_d]$) are obtained by unnormalization using following equations:
        \begin{equation}
            \small
            \tilde{a}_{d} = \tilde{a}_{n,d}f_x+c_x,\tilde{b}_{d} = \tilde{b}_{n,d}f_y+c_y.
            \label{eq:hb_4}
        \end{equation}
        
        We apply this process to all joints $J$ and frames $T$ to obtain a distorted 2D trajectory (denoted as $\tilde{\mathbf{P}}_{dist}$) reflecting a specific distortion. Then, we consider a pair of the distorted 2D trajectory and ground-truth 3D joints (denoted as $\mathbf{s} = [\mathbf{x},\mathbf{y},\mathbf{z}]\in\mathbb{R}^{J\times3}$) as a single task $\mathcal{T}$ of MAML.
        
        This method is highly efficient because it does not apply distortion in image domain, and thus, we can generate numerous distortions in the training phase, as shown in Figure~\ref{fig:overall} (a). Furthermore, it can reflect the jittered outputs of the 2D keypoint detector caused by inherent ambiguity (\eg, occlusion), because it generates distorted joints from predicted 2D keypoints. Synthetic tasks can also be generated from ground-truth 3D joints, as shown in Figure~\ref{fig:synthetic data} (b). In this case, normalized 2D keypoints are obtained from the ground-truth 3D joints through projection. However, as shown in Table~\ref{tbl:synthetic2Dgen}, this method is less effective because it could not reflect the noisy output of the 2D keypoint detector, resulting in a domain gap during training and testing.
        
    \subsection{Training Phase}
        \label{sec:Training Phase}
        In the training phase, we will perform meta-learning using synthetic distorted tasks from undistorted videos. Our goal in the training phase is to find sensitive initial transferable weights to camera distortion by utilizing optimization-based meta-learning. Our algorithm mostly follows the framework of MAML, but to achieve better performance, there are two modifications: \textit{stratified sampling} and \textit{random distortion pretraining}.
    
        As shown in Figure~\ref{fig:overall} (a), given predicted undistorted 2D trajectory, we generate a batch of distorted 2D trajectories (denoted as $\{\tilde{\mathbf{P}}_{dist,i}\}_{i=1}^N$, where $N$ represents the number of tasks in meta-batch) with sampled distortion parameters. Then, we construct each task $\mathcal{T}_i$ by pairing a distorted 2D trajectory $\tilde{\mathbf{P}}_{dist,i}$ and ground-truth 3D joints $\mathbf{s}$.
        
        Specifically, parameters related to radial distortion are sampled by $k_1,k_2,k_3 \sim \mathcal{U}[-\lambda_1,\lambda_1]$ and tangential distortion parameters are sampled by $p_1,p_2 \sim \mathcal{U}[-\lambda_2,\lambda_2]$ where $\lambda_1$ and $\lambda_2$ denote the maximum value of each distribution. We basically use the sampling method to both task-level training and task-level testing. However, for the task-level training, we adopt \textit{stratified sampling} for sampling parameter $k_1$, which determines the main form of distortion. In this case, a $k_1$ of $i$th sample in the meta-batch is sampled as follows:
        \vspace{-1mm}
        \begin{equation}
            k_{1,i} \sim -\lambda_1 + 2\cdot\lambda_1\cdot\mathcal{U}\left[ \frac{i-1}{N},\,\, \frac{i}{N} \right]\,.
            \label{eq:stratified sampling}
            \vspace{-1mm}
        \end{equation}
        
        By sampling the distortion parameter $k_1$ from evenly spaced bins, the meta-batch can consist of tasks with varying degrees of distortion. This enhances the adaptability of our network as shown in Table~\ref{tbl:effectiveness of each method}. We denote the distribution of tasks generated using stratified sampling as $p_{strat}(\mathcal{T})$, and using only uniform distribution as $p_{rand}(\mathcal{T})$.
        
        Finally, we consider a 3D pose estimator model represented by a parameterized function $g_\theta$ with parameters $\theta$. We perform only one gradient descent update when the parameters $\theta$ is adapted to a new task $\mathcal{T}_i$. Thus, the newly adapted parameters $\theta_i'$ are obtained by
        \vspace{-1mm}
        \begin{equation}
            \theta_i' = \theta - \alpha \nabla_{\theta} \mathcal{L}_{\mathcal{T}_{i}}(g_{\theta}),
            \vspace{-1mm}
        \end{equation}
        where $\alpha$ is the learning rate for task-level training.
        
        The parameters $\theta$ of model are optimized by maximizing the performance of $g_{\theta_i'}$ with respect to $\theta$ across tasks sampled for task-level testing. Specifically, the meta-objective is expressed as follows:
        \vspace{-1mm}
        \begin{equation}
            \begin{aligned}
                &\arg \min_{\theta} \sum_{\mathcal{T}_i\sim p(\mathcal{T})} \mathcal{L}_{\mathcal{T}_{i}}(g_{\theta_i'})\\
                =&\arg \min_{\theta} \sum_{\mathcal{T}_i\sim p(\mathcal{T})} \mathcal{L}_{\mathcal{T}_{i}}(g_{\theta - \alpha \nabla_{\theta} \mathcal{L}_{\mathcal{T}_{i}}(g_{\theta})}).
                \label{eq:meta-objective}
            \end{aligned}
            \vspace{-1mm}
        \end{equation}
        
        Finally, we perform meta-optimization by using the Eq.~\ref{eq:meta-objective}. For the stochastic gradient descent, model parameters $\theta$ are updated as follows:
        \vspace{-1mm}
        \begin{equation}
            \theta \leftarrow \theta - \beta \nabla_{\theta} \sum_{\mathcal{T}_{i}\sim p(\mathcal{T})} \mathcal{L}_{\mathcal{T}_i} (g_{\theta_i'}),
            \label{eq:meta-optimization}
            \vspace{-1mm}
        \end{equation}
        where $\beta$ is the learning rate for meta-optimization. We use a loss function, MPJPE, which is the L2 distance between ground-truth 3D joints and predicted ones as a task-level objective in the entire process of meta-optimization.
        
        Additionally, we pretrain the network before training meta-learner through \textit{random distortion pretraining} that regresses 3D joints from randomly distorted 2D keypoint trajectories. This allows the network to learn feature representation under various distortions and consequently enables stable MAML training. However, while random distortion pretraining helps in the stability of MAML, the pretraining \emph{without} meta-learning shows poor results, as shown in Figure~\ref{fig:fast_adaptation}, when the network is adapted to the specific distortion, because it is not a transferable initial weights.
    
    \subsection{Adaptation before Testing}
        When the trained model that can quickly adapt to arbitrary distortions is used, it must, first, be adapted to the specific distortion of the testing environment. This is similar to meta-testing in the MAML framework. We assume two practical situations, \textit{Scenario 1} and \textit{Scenario 2}, and propose an adaptation method for each case.
        \vspace{-4mm}
        
        \paragraph{Scenario 1} is a situation in which a user can collect data using motion capture sensors in front of a testing environment. In this case, data with the same distortion as the testing environment can be obtained. Thus, we adopt naive fine-tuning using the MPJPE loss function to adapt the network to the specific distortion, as shown in Figure~\ref{fig:overall} (b). The collected data would be in relatively small amounts than the large-scale dataset (\eg, Human3.6M). Therefore, we will check whether the network can adapt well with small amounts of collected data. Detailed settings are provided in Section~\ref{exp:datasets and evaluation}.
        \vspace{-4mm}
        
        \paragraph{Scenario 2} is a situation when the user is unable to obtain data in a testing environment. In this case, the network should be adapted to the specific distortion using only test videos. As shown in Figure~\ref{fig:overall} (c), we adopt Inference Stage Optimization (ISO)~\cite{ref18_ISO_NeurIPS2020}, which performs self-supervised training using the test data before testing. Usually, the inferred 3D joints are orthogonally projected to 2D plane and compared with the predicted 2D keypoints to perform ISO. However, if there is distortion in the video, this method cannot be used. Therefore, we propose the novel ISO method which utilizes \textit{bone-length symmetry} and \textit{bone-length consistency} that allows self-supervision within the inferred 3D joint itself (details in Appendix ~\ref{supp:bone-length based ISO}). The former constrains the length of a person's left and right bones to be equal, whereas the latter constrains each bone to be equal in length between consecutive frames within a video. The constraints based on bone-length have been used for regularization in fully-supervised training, but have never been used for ISO. Also, these methods might be a weak constraint, but our network is fully adaptable even with these constraints because it has been sensitively trained on distortion via MAML.
    
    \subsection{Algorithm}
        \vspace{-1mm}
        Algorithm \ref{alg:1} shows the entire process of Section \ref{sec:Training Phase}. As shown in lines 2-8, random distortion pretraining is performed before meta-learning. Subsequently, meta-learning is performed, as shown in lines 9-17. Lines 11-14 and lines 15-16 present task-level training and meta-optimization with task-level testing, respectively.
        \vspace{-1mm}

\section{Experiments}
\subsection{Datasets and Evaluation}
    \label{exp:datasets and evaluation}
\begin{figure}[t]
    	\begin{center}
    		\captionsetup{justification=centering}
    		\begin{subfigure}[t]{0.24\linewidth}
    			\centering
    			\includegraphics[width=1\columnwidth]{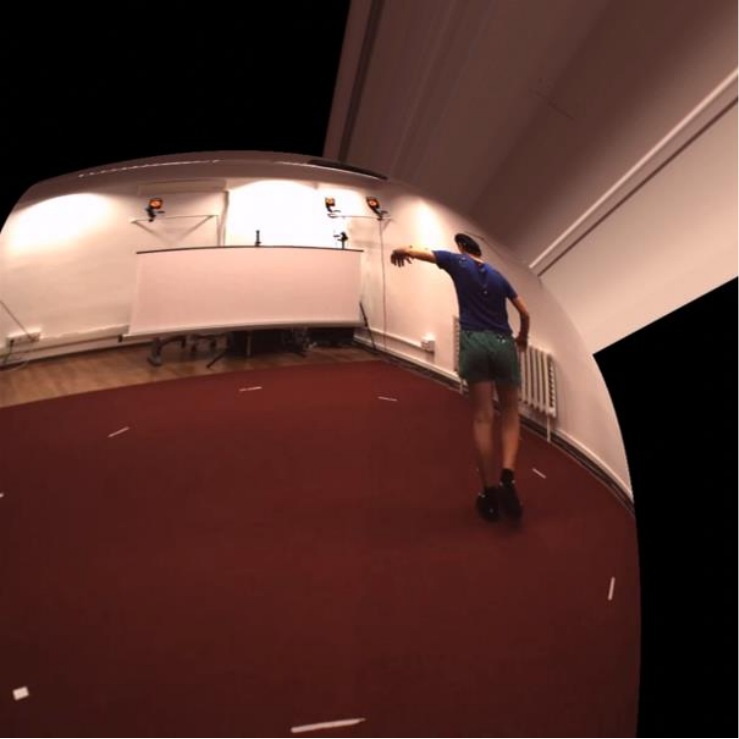}
    			\caption{B+T ($d_1$)}
    		\end{subfigure}
    		\begin{subfigure}[t]{0.24\linewidth}
    			\centering
    			\includegraphics[width=1\columnwidth]{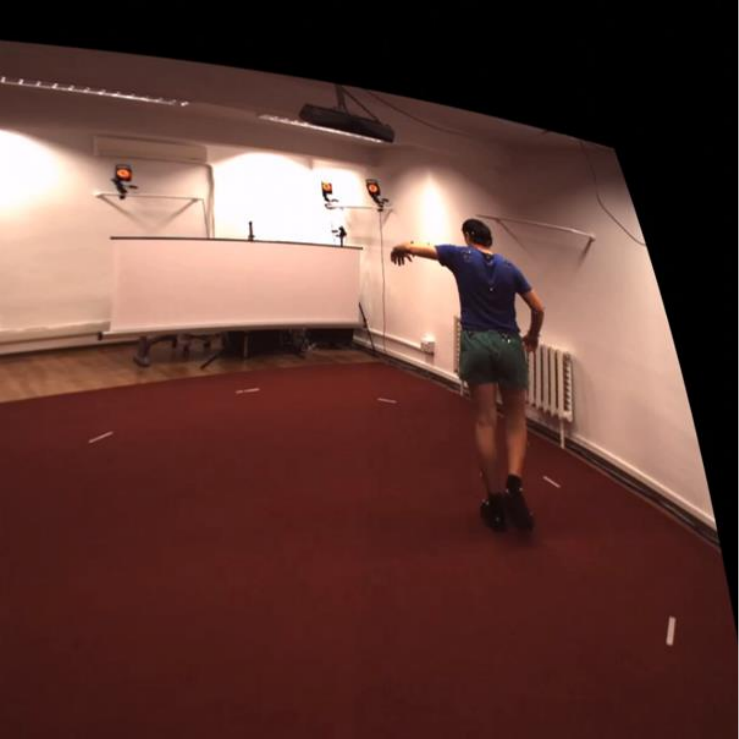}
    			\caption{P+T ($d_2$)}
    		\end{subfigure}
    		\begin{subfigure}[t]{0.24\linewidth}
    			\centering
    			\includegraphics[width=1\columnwidth]{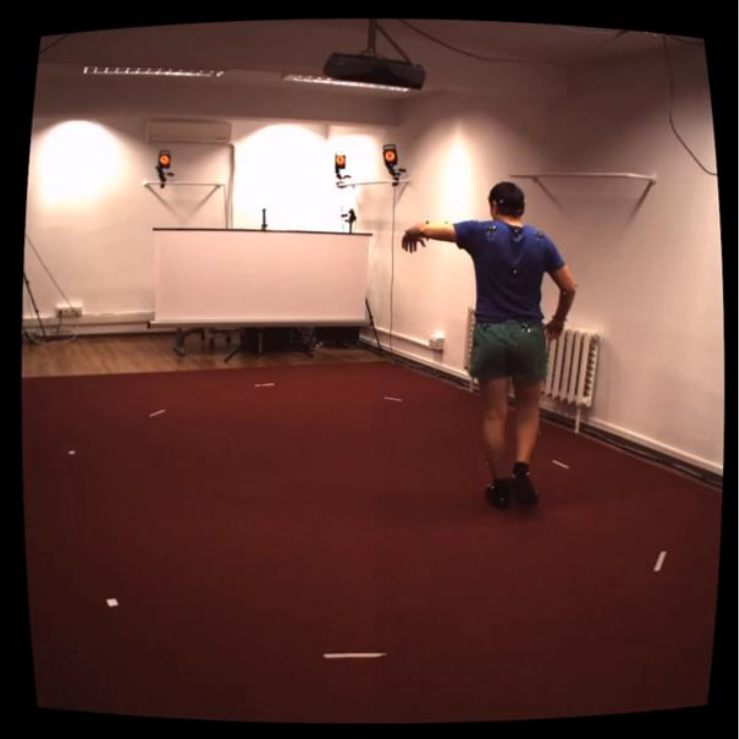}
    			\caption{B+T ($d_3$)}
    		\end{subfigure}
    		\begin{subfigure}[t]{0.24\linewidth}
    			\centering
    			\includegraphics[width=1\columnwidth]{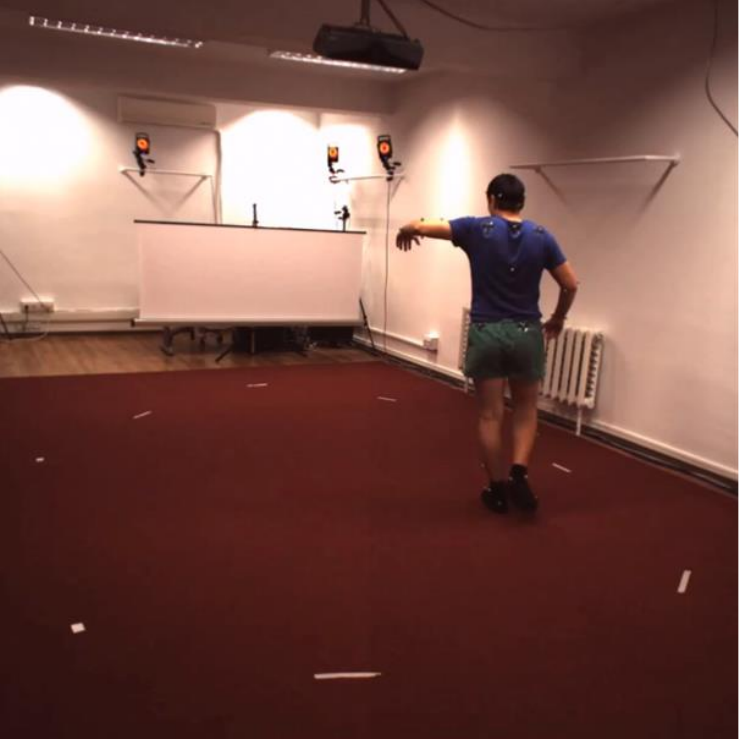}
    			\caption{P+T ($d_4$)}
    		\end{subfigure}
    		\vspace{-1mm}
    	\end{center}
    	    \vspace{-5mm}
    		\caption{Videos rendered with different kinds of distortion. B, P, and T represent barrel, pincushion, and tangential distortion respectively. For (a) and (b) heavy distortion is applied, and moderate distortion is applied to (c) and (d).}
    		\vspace{-4mm}
    	\label{fig:dataset_distorted}
    \end{figure}

    \paragraph{Human3.6M~\cite{ref19_6682899}} is a large-scale dataset containing 3.6 million video frames and corresponding 2D and 3D human keypoint labels. We construct a cross-scenario on distortion to validate the effectiveness of the proposed method. For training, we use five subjects (S1, S5, S6, S7, S8) with undistorted videos, as in previous works~\cite{ref9_pavllo20193d,ref11_anatomy3D}, since our method can generate synthetic distorted tasks from undistorted 2D keypoints. For testing, only one subject (S11) is used. Due to the absence of test videos with distortion, we generate four different kinds of distorted videos (denoted as $d_1$, $d_2$, $d_3$, and $d_4$, details in Appendix ~\ref{supp:about distortion params}) from undistorted videos of S11 by using Blender\footnote{\url{https://www.blender.org/}} software, as shown in Figure~\ref{fig:dataset_distorted}. We evaluate the proposed method in each kind of distortion. For \textit{Scenario 1} in adaptation before the testing phase, collected small-scale dataset with the same distortion as the test data is required. Therefore, we adopt only 1\% of S9 and apply the same distortion as S11 to it.
    \vspace{-3mm}
    
    \paragraph{Evaluation Metrics.}
        We use three evaluation protocols following previous works~\cite{ref7_martinez2017simple,ref8_zhao2019semantic,ref9_pavllo20193d,ref11_anatomy3D,ref12_liu2020attention,ref18_ISO_NeurIPS2020}. The first is mean per joint position error (MPJPE) in millimeters, the L2 distance between the predicted 3D joints and ground-truth joints. The second is P-MPJPE. This is similar to MPJPE, but calculates the error between the joints after alignment using Procrustes Analysis. The last one is percentage of correct 3D joints with a threshold as 50\% of the head segment length (PCKh@0.5).

    \begin{algorithm}[t]
    	\DontPrintSemicolon
    	\SetAlgoLined
    	\KwInput{$\mathcal{{D}}$: a large-scale 3D human pose dataset}
    	\KwInput{$\alpha$, $\beta$: learning rate hyperparameters}
    	\KwOutput{Model parameters $\theta$}
    	Randomly initialize $\theta$\\
    	\While{not done}
    	{
    	    Sample batch of tasks $\mathcal{T}_{rand,i} \sim p_{rand}(\mathcal{T})$\\
    		\For{all $\mathcal{T}_{rand,i}$}{
    			Calculate loss by MPJPE: $\mathcal{L}_{\mathcal{T}_{rand,i}}(g_{\theta})$\\
    			Compute updated parameters: $\theta=\theta - \beta \nabla_{\theta} \mathcal{L}_{\mathcal{T}_{rand,i}}(g_{\theta})$\\
    		}
    	}
    	\While{not done}
    	{
    		Sample batch of tasks $\mathcal{T}_{strat,i} \sim p_{strat}(\mathcal{T})$\\
    		\For{all $\mathcal{T}_{strat,i}$}{
    		    Calculate loss by MPJPE: $\mathcal{L}_{\mathcal{T}_{strat,i}}(g_{\theta})$\\
    			Compute updated parameters: $\theta_i'=\theta - \alpha \nabla_{\theta} \mathcal{L}_{\mathcal{T}_{strat,i}}(g_{\theta})$
    		}
    		Update $\theta$ with respect to average test loss:\\
    		$\theta \leftarrow \theta - \beta \nabla_{\theta} \sum_{\mathcal{T}_{rand,i}\sim p_{rand}(\mathcal{T})} \mathcal{L}_{\mathcal{T}_{rand,i}}(g_{\theta_i'})$
    		
    	}
    	\caption{Training Phase}
    	\label{alg:1}
    \end{algorithm}

    \begin{table*}[t]
        \centering
        \small
        \tabcolsep=1mm
        \resizebox{0.83\textwidth}{!}{
            \begin{tabular}{@{}l|ccc|ccc@{}}
            \toprule
            & & \textit{Scenario 1} & & & \textit{Scenario 2} & \\
            \cmidrule{2-7}
            Method & MPJPE($\downarrow$) & P-MPJPE($\downarrow$) & PCKh@0.5($\uparrow$) & MPJPE($\downarrow$) & P-MPJPE($\downarrow$) & PCKh@0.5($\uparrow$)\\
            \midrule
            Martinez \etal \cite{ref7_martinez2017simple} ICCV'17   &~~ \underline{78.3} / 63.1 ~~&~~ \underline{58.1} / 48.7 ~~&~~ 66.6 / 76.5 ~~&~~ 128.0 / 68.3 ~~&~~ 86.8 / 49.1 ~~&~~ 47.3 / 74.1 ~~ \\
            Zhao \etal \cite{ref8_zhao2019semantic} CVPR'19         &~~ 86.3 / 64.0 ~~&~~ 64.2 / 47.4 ~~&~~ 63.2 / 76.9 ~~&~~ 119.7 / 71.4 ~~&~~ 85.5 / 51.9 ~~&~~ 45.0 / 72.2 ~~ \\
            \midrule
            Pavllo \etal \cite{ref9_pavllo20193d} CVPR'19           &~~ 79.9 / 65.0 ~~&~~ 59.4 / 48.3 ~~&~~ \underline{67.3} / 76.7 ~~&~~ 114.1 / 64.5 ~~&~~ 72.4 / \underline{45.7} ~~&~~ 47.9 / 76.6 ~~ \\
            Chen \etal \cite{ref11_anatomy3D} TCSVT'21              &~~ 89.4 / \underline{62.7} ~~&~~ 61.9 / \underline{46.3} ~~&~~ 59.2 / \underline{77.8} ~~&~~ \underline{107.3} / 65.1 ~~&~~ \underline{71.0} / 46.3 ~~&~~ 49.0 / \underline{77.3} ~~ \\
            Liu \etal \cite{ref12_liu2020attention} CVPR'20         &~~ 81.5 / 68.8 ~~&~~ 60.9 / 51.0 ~~&~~ 66.4 / 74.7 ~~&~~ 110.7 / \underline{64.0} ~~&~~ 77.5 / 46.5 ~~&~~ \underline{49.5} / 76.8 ~~ \\
            \midrule
            Ours                                                    &~~ \textbf{62.0} / \textbf{53.6} ~~&~~ \textbf{46.4} / \textbf{40.6} ~~&~~ \textbf{78.4} / \textbf{83.3} ~~&~~ \textbf{66.1} / \textbf{51.6} ~~&~~ \textbf{47.8} / \textbf{39.2} ~~&~~ \textbf{76.3} / \textbf{85.7} ~~ \\
            \bottomrule
            \end{tabular}
        }
    	\caption{Comparison of average performance on (heavy) / (moderate) with other state-of-the-art models. The top two rows~\cite{ref7_martinez2017simple,ref8_zhao2019semantic} are based on a single-frame and others~\cite{ref9_pavllo20193d,ref11_anatomy3D,ref12_liu2020attention}, including our method, are based on a video with a frame length of 27. Best in bold, second-best underlined. More results can be seen in the supplementary material (Appendix ~\ref{supp:quantitative results}).}
        \label{tbl:sota}
    \end{table*}

    \begin{figure*}[t]
        \begin{center}
            \includegraphics[width=1\linewidth]{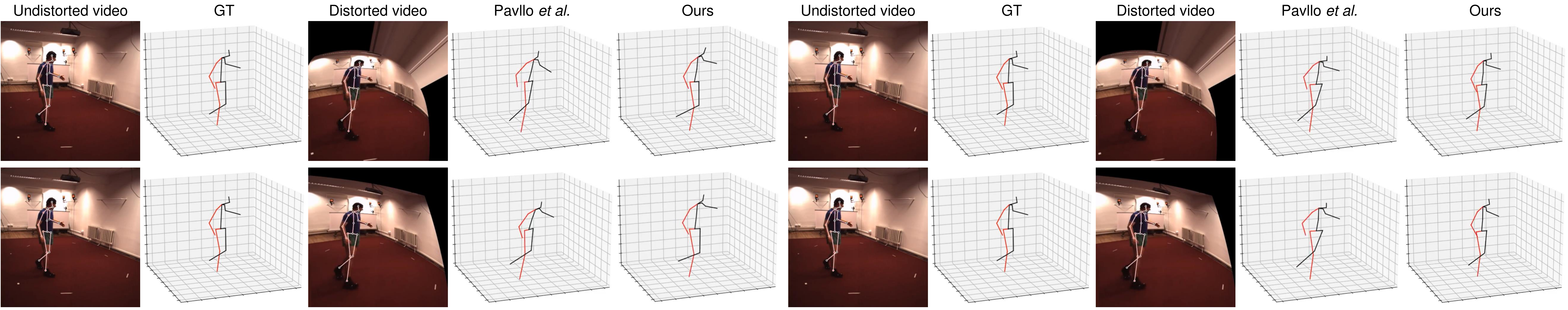}
        \end{center}
        \vspace{-3mm}
        \caption{Qualitative results on heavily distorted videos of Human3.6M. The five columns from the leftmost are the result under the \small\textit{Scenario 1} \normalsize setting, while the rest columns are the result under the \small\textit{Scenario2} \normalsize setting. \textbf{Top row:} 3D reconstruction results on $d_1$. \textbf{Bottom row:} 3D reconstruction results on $d_2$. More results can be seen in Appendix ~\ref{supp:qualitative results}.}
        \vspace{-1mm}
        \label{fig:qualitative}
        \vspace{-3mm}
    \end{figure*}

    \subsection{Implementation Details}
        \label{sec:implementation details}
        \vspace{-1mm}
        For 3D pose estimator, the proposed method is not about network architecture but training methods. Therefore, we adopt the state-of-the-art model for 3D human pose estimation in video proposed in the previous work~\cite{ref9_pavllo20193d} as our base model. It is fully convolutional and based on dilated temporal convolutions with residual blocks. For 2D keypoint detector, we use \textit{Mask R-CNN}~\cite{ref14_He_2017_ICCV} with a ResNet-101-FPN~\cite{ref35_FPN} backbone as off-the-shelf 2D keypoint detector. We fine-tune the COCO~\cite{ref34_mscoco} pretrained model on 2D keypoints of Human3.6M. Similar to previous work~\cite{ref9_pavllo20193d}, the 2D keypoint format of COCO differs from Human3.6M, and hence, we reinitialize the last layer of the keypoint network of the detector and carry out fine-tuning, after which the 2D keypoint detector is frozen in the entire training process for the 3D pose estimator because it is robust to camera distortion. We use Adam~\cite{ref36_Adam} optimizer, with batch size $1024$. During the training phase, we set the required $\lambda_1$ and $\lambda_2$ to 5 and 0.5 respectively, to sample the distortion parameters. The learning rate $\alpha$ for task-level training is set to $0.1$ and the $\beta$ for meta-optimization is set to $0.001$. We use $5$ for the number of samples in meta-batch. The learning rate decay is set to $0.95$ and network is trained with $60$ epochs. In both \textit{Scenario 1} and \textit{Scenario 2} of adaptation before testing, learning rate is set to $0.6$ and epochs for adaptation is set to $100$. Note that, during the adaptation process, we train the model for $100$ epochs, however since it is done on a small-scale dataset, the overall training time required for the adaptation is within a few minutes.
    
    \subsection{Experiment Results}
        \label{sec:experiment results}
        \vspace{-1mm}
        In this section, we validate effectiveness of the proposed method. We evaluate performance on four different kinds of distortion in all experiments and report the average performance of $d_1$ and $d_2$ which applied heavy distortion and the average performance of $d_3$ and $d_4$ which applied moderate distortion. All reported values are performance after adaptation to the specific distortion. For \textit{Scenario 1}, the trained network is adapted by fine-tuning on 1\% of S9, which went through the same distortion as S11. For \textit{Scenario 2}, the network is adapted by ISO on 0.1\% of test videos (S11).
        \vspace{-3mm}
        \vspace{-2mm}
        \paragraph{Comparison with State-of-the-Art.}
            Table~\ref{tbl:sota} shows the performance of existing 3D pose estimation algorithms and our method. The baseline models do not take cross-scenario about distortion into account. However, for fair evaluation, we evaluate the performance in \textit{Scenario 1} after fine-tuning on the small-scale dataset, just like our method. For \textit{Scenario 2}, we did not apply ISO to baseline models because they show poor performance when ISO is applied.
        
            The proposed method outperforms other methods regardless of the kinds of distortions and scenarios. Specifically, compared to our base model~\cite{ref9_pavllo20193d}, the proposed method shows -14.64mm, -10.35mm, and +17.7\% average performance improvement in \textit{Scenario 1} for each metric (\ie, MPJPE, P-MPJPE, and PCKh@0.5) and -30.45mm, -15.55mm, and +18.75\% average performance improvement in \textit{Scenario 2}. Especially in \textit{Scenario 2}, our method rather shows better performance than \textit{Scenario 1} under moderate distortion. This demonstrates that our bone-length-based ISO method is effective and also that the trained model has transferable initial weights.
        
            Figure~\ref{fig:qualitative} shows qualitative results of the estimated 3D pose from distorted videos. Unlike others, our method successfully adapts to the distortion that the test video has, and consequently, we can see that estimated 3D joints from the distorted video by using the proposed method is almost the same with ground-truth joints. Results predicted from videos with more diverse poses and distortion can be seen in the supplementary material (Appendix ~\ref{supp:qualitative results}).

        \begin{table}[t]
            \centering
        	\small
        	\tabcolsep=1mm
        	\resizebox{\columnwidth}{!}{
        		\begin{tabular}{@{}l|ccc@{}}
        		\toprule
        		& MPJPE($\downarrow$) & P-MPJPE($\downarrow$) & PCKh@0.5($\uparrow$) \\
        		\midrule
        		\textit{base model} \cite{ref9_pavllo20193d}    &~~ 84.2 / 79.6 ~~&~~ 62.8 / 59.7 ~~&~~ 64.8 / 66.9 ~~ \\
        		+ MAML (with synthetic tasks)                   &~~ 73.5 / 67.5 ~~&~~ 55.6 / 51.7 ~~&~~ 72.0 / 74.5 ~~ \\
        		+ \textit{stratified sampling}                   &~~ 71.7 / 66.2 ~~&~~ 54.3 / 50.4 ~~&~~ 72.8 / 75.2 ~~ \\
        		+ \textit{random distortion pretraining}         &~~ 67.2 / 61.9 ~~&~~ 51.0 / 47.0 ~~&~~ 75.7 / 78.2 ~~ \\
        		\bottomrule
        		\end{tabular}
        	}
        	\caption{Effectiveness of each proposed method based on input frame length of 9 under \small\textit{Scenario 1} \normalsize setting. Each value denotes performance on (distortion $d_1$) / (distortion $d_2$).}
        	\vspace{-1mm}
        \label{tbl:effectiveness of each method}
        \end{table}

        \begin{table}[t]
            \centering
        	\small
        	\tabcolsep=1mm
        	\resizebox{\columnwidth}{!}{
        		\begin{tabular}{@{}l|ccc@{}}
        		\toprule
        		Method& MPJPE($\downarrow$) & P-MPJPE($\downarrow$) & PCKh@0.5($\uparrow$) \\
        		\midrule
                Predicted 2D keypoints     &~~ 62.0 / 53.6 ~~&~~ 46.4 / 40.6 ~~&~~ 78.4 / 83.3 ~~ \\
                Ground-truth 3D joints     &~~ 64.7 / 56.1 ~~&~~ 48.2 / 42.0 ~~&~~ 77.0 / 82.0 ~~ \\
                \midrule
                Predicted 2D keypoints     &~~ 66.1 / 51.6 ~~&~~ 47.8 / 39.2 ~~&~~ 76.3 / 85.7 ~~ \\
                Ground-truth 3D joints     &~~ 71.3 / 55.6 ~~&~~ 51.9 / 42.6 ~~&~~ 72.8 / 83.5 ~~ \\
        		\bottomrule
        		\end{tabular}
        	}
        	\caption{Comparison of average performance on (heavy) / (moderate) between the methods generating synthetic 2D keypoints. \textbf{Top rows:} \small\textit{Scenario 1}\normalsize. \textbf{Bottom rows:} \small\textit{Scenario 2}\normalsize.}
        	\vspace{-2mm}
        	\vspace{-2mm}
        \label{tbl:synthetic2Dgen}
        \end{table}

        \vspace{-4mm}
        \paragraph{Ablation Studies.}
            We first look at the contribution of each of the proposed method. We evaluate the performance changes, adding each proposed method with Pavllo \etal \cite{ref9_pavllo20193d} as our base model. As shown in Table~\ref{tbl:effectiveness of each method}, we can notice that each method provides a positive contribution under all metrics. In particular, the significant improvement comes from utilizing MAML using synthetic distorted tasks and learning rich feature representation on distortion through random distortion pretraining.
            
            Table~\ref{tbl:synthetic2Dgen} shows the performance when applying each of the two methods that generate synthetic distorted 2D keypoints based on frame length of 27. Predicted 2D keypoints denotes the case in which synthetic distorted 2D keypoints are generated using noisy results inferred from the 2D keypoint detector, as shown in Figure~\ref{fig:synthetic data} (a) and Ground-truth 3D keypoints denotes the case in which synthetic keypoints are generated using the ground-truth 3D keypoints as shown in Figure~\ref{fig:synthetic data} (b). We can notice that the former method shows better performance under all metrics and scenarios since there is less domain gap between training and testing.
            
            Table~\ref{tbl:performance and complexity} reports the performance and complexity of the model (\ie, parameters and FLOPs) with respect to different input frame lengths. Our method uses the same model structure as Pavllo \etal \cite{ref9_pavllo20193d} because our work is about the learning method rather than the structure of the model. When the input frame length is 3, the proposed method shows comparable performance to the base model, even though the capacity is one-fifty of the base model. Furthermore, our method with an input frame length of 27 outperforms the base model of the same size significantly. Moreover, our method has the same floating-point operations (FLOPs) for inference as the base model~\cite{ref9_pavllo20193d}, thus no additional computational cost is required compared to the base model when testing after adaptation to the test environment.

        \begin{table}[t]
        	\centering
        	\small
        	\tabcolsep=1mm
        	\resizebox{\columnwidth}{!}{
        		\begin{tabular}{@{}l|r|r|r|r|r@{}}
        		\toprule
        		Model & ~Parameters~ & ~$\approx$ FLOPs~ & ~MPJPE~ & ~P-MPJPE~ & ~PCKh@0.5~ \\
        		\midrule
        		Pavllo \etal \cite{ref9_pavllo20193d} 27f~ & 8.56M & 17.11M & 72.4 & 53.8 & 72.0 \\
        		\midrule
                Ours 3f     & 0.16M & 0.32M & 75.0 & 56.1 & 69.6 \\
                Ours 9f     & 4.36M & 8.71M & 59.8 & 45.4 & 79.5 \\
                Ours 27f    & 8.56M & 17.11M & 57.6 & 43.4 & 80.9 \\
        		\bottomrule
        		\end{tabular}
        	}
        	\caption{Performance and computational complexity of various models under \small\textit{Scenario 1}\normalsize. The reported performance is the average value for all kinds of distortions.}
        	\label{tbl:performance and complexity}
        	\vspace{-2mm}
        \end{table}
        
        \begin{figure}[t]
            \begin{center}
                \includegraphics[width=1\linewidth]{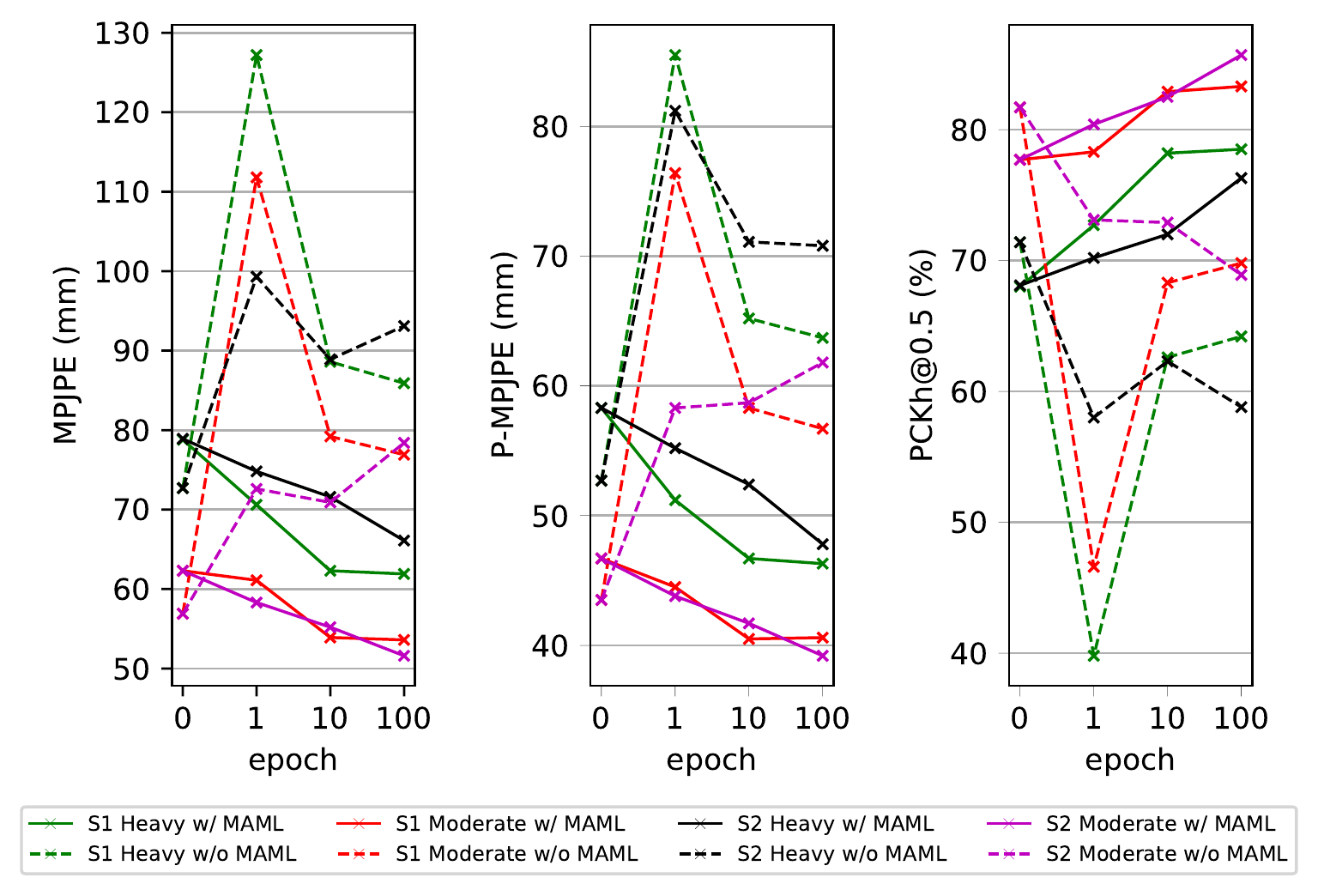}
            \end{center}
            \vspace{-6mm}
            \caption{Performance changes during adaptation to the specific distortion. S1 and S2 denote \small\textit{Scenario 1 }\normalsize and \small\textit{Scenario 2}\normalsize, respectively. A solid line w/ MAML denotes our final model trained using all the elements proposed in Section~\ref{sec:Training Phase}, and a dashed line w/o MAML denotes a model trained only with \textit{random distortion pretraining}.}
            \vspace{-1mm}
            \label{fig:fast_adaptation}
            \vspace{-2mm}
            \vspace{-2mm}
        \end{figure}

        \vspace{-4mm}
        \paragraph{Performance Changes during Adaptation.}
            We also validate the ability of the model, trained in the training phase, to adapt well to the specific camera distortion. In this experiment, we observe the performance changes of the model with and without MAML during the adaptation process based on frame length 27. As shown in Figure~\ref{fig:fast_adaptation}, in the case of a model using MAML, we can notice that it adapts well regardless of the degrees of distortion and scenarios. Also, the mean and standard deviation of MPJPE are 6.5mm (10\%) and 2.3mm (25\%) lower than those of w/o MAML (at epoch 0), respectively. In contrast, in the case of the model not using MAML, the model is not stably adapted, and its performance is rather significantly degraded. Specifically, it performs well at epoch $0$ with the effect of the proposed random distortion pretraining, however since it is not a transferable initial weight, it is highly degraded when the adaptation process starts. This demonstrates the superior potential of MAML to adapt to various distortion environments. Note that, as mentioned in Section~\ref{sec:implementation details}, training time required for the adaptation process is within a few minutes.
\vspace{-6mm}

\section{Conclusion}
    \vspace{-2mm}
    We have introduced a model for 3D human pose estimation that can adapt quickly to arbitrary camera distortion. Our model finds initial transferable weights that are sensitive to distortion through meta-learning. For this, we overcome the limitations of the absence of publically available distorted data by generating synthetic distorted tasks from undistorted data. Furthermore, we propose a novel ISO method based on bone-length that can adapt the model to the test environment without 3D joint labels. Our method is expected to be very useful in practice because once trained, it can adapt to any distortion without camera calibration.

%% file: supplementary.tex
\clearpage
\newpage
\appendix

\section{Supplementary material}
    \subsection{About Distortion Parameters}
        \label{supp:about distortion params}
        Our work targets two main types of in-the-wild situations. The first is the distortion that occurs in cameras for special purposes, such as fisheye and wide-angle cameras (\eg, insta360), and the second is the distortion that occurs in low-cost cameras such as surveillance cameras. We defined the former and latter as ``heavy'' and ``moderate'' (equivalent to ``light'') distortion, respectively. Since there is no common benchmark and public dataset with such level of distortions, we randomly selected two sets of parameters (\ie, $k_1,k_2,k_3,p_1,p_2$) that well reflect real-world situations at each distortion, and we synthesized videos and used them for evaluation. The $d_1$, $d_2$, $d_3$, and $d_4$ have distortion parameters of \small($\mp4.142$, $\pm4.956$, $\mp0.062$, -$0.488$, -$0.712$)\normalsize, \small($\mp2.071$, $\pm2.478$, $\mp0.031$, -$0.010$, -$0.014$)\normalsize, respectively. The original distortion present in H3.6M is \small(-$0.207$, $0.248$, -$0.003$, -$0.001$, -$0.001$)\normalsize, which is almost identical to no distortion.
    
    \subsection{Bone-Length based ISO}
        \label{supp:bone-length based ISO}
        Given a video clip with frame length of $T$, predicted 3D joints $\tilde{\mathbf{S}} = \{\tilde{\mathbf{s}}_t\}_{t=1}^T \in\mathbb{R}^{T\times J\times3}$ where $\tilde{\mathbf{s}}_t\in\mathbb{R}^{J\times3}$ represents the predicted 3D joints at frame $t$ can be obtained. Then, we can get the predicted bone-lengths (denoted as $\tilde{\mathbf{l}} = \{\tilde{l}_{t,j}\}_{t=1}^T\in\mathbb{R}^{T\times (J-1)}$ where $\tilde{l}_{t,j}$ denotes the predicted length of $j$th bone at frame $t$) from the predicted 3D joints by calculating the distance between adjacent joints. Finally, we can calculated the \textit{bone-length symmetry} loss as follows:
        \vspace{-1mm}
        \begin{equation}
            \mathcal{L_{\text{symmetry}}} = \sum_{t=1}^{T}\sum_{(j_l,j_r)\in\mathcal{P}}\abs{\tilde{l}_{t,j_l}-\tilde{l}_{t,j_r}},
            \vspace{1mm}
        \end{equation}
        where $\mathcal{P}$ contains all the pair of bones that are symmetrical to the left and right (denoted as $j_l$ and $j_r$, respectively). Also, the \textit{bone-length consistency} loss is obtained by:
        \begin{equation}
            \mathcal{L_{\text{consistency}}} = \sum_{t=1}^{T-1}\sum_{j=1}^{J-1}\abs{\tilde{l}_{t+1,j}-\tilde{l}_{t,j}}.
        \end{equation}
        Thus, our final objective for the Inference Stage Optimization in \textit{Scenario 2} is as follows:
        \begin{equation}
            \mathcal{L_{\text{ISO}}} = \mathcal{L_{\text{symmetry}}} + \mathcal{L_{\text{consistency}}}.
        \end{equation}

    \subsection{Quantitative Results}
        \label{supp:quantitative results}
        In Table~\ref{tbl:sota}, we provided the average performance on each of the heavy distortion and moderate distortion. Table~\ref{tbl:supp_quantitative_all_distortion} shows the performance at each distortion (\ie, $d_1$, $d_2$, $d_3$, and $d_4$). In addition, Table~\ref{tbl:supp_quantitative_actionwise} shows the reconstruction accuracy (PCKh@0.5) for each action. The reported accuracy here are the average value for all kinds of distortions. We can notice that our method outperforms other methods regardless of the kinds of distortions and actions.

    \begin{table*}[ht]
        \begin{subtable}{\linewidth}
            \centering
            \small
            \tabcolsep=1mm
            \resizebox{0.83\textwidth}{!}{
                \begin{tabular}{@{}l|ccc|ccc@{}}
                \toprule
                & & \textit{Scenario 1} & & & \textit{Scenario 2} & \\
                \cmidrule{2-7}
                Method & MPJPE($\downarrow$) & P-MPJPE($\downarrow$) & PCKh@0.5($\uparrow$) & MPJPE($\downarrow$) & P-MPJPE($\downarrow$) & PCKh@0.5($\uparrow$)\\
                \midrule
                Martinez \etal \cite{ref7_martinez2017simple} ICCV'17   &~~ \underline{81.1} / \underline{75.5} ~~&~~ \underline{59.6} / \underline{56.6} ~~&~~	65.4 / 67.8	~~&~~ \underline{92.8} / 163.2 ~~&~~ \underline{65.2} / 108.3 ~~&~~ 58.3 / 36.2 ~~ \\
                Zhao \etal \cite{ref8_zhao2019semantic} CVPR'19         &~~ 90.7 / 81.8	~~&~~ 66.0 / 62.3 ~~&~~	61.3 / 65.1	~~&~~ 104.6 / 134.7 ~~&~~ 76.0 / 94.9 ~~&~~ 52.6 / 37.3 ~~ \\
                \midrule
                Pavllo \etal \cite{ref9_pavllo20193d} CVPR'19           &~~ 83.2 / 76.6	~~&~~ 61.4 / 57.3 ~~&~~	\underline{65.5} / \underline{69.1}	~~&~~ 94.4 / 133.8 ~~&~~ 65.6 / 79.2 ~~&~~ 57.5 / 38.2 ~~ \\
                Chen \etal \cite{ref11_anatomy3D} TCSVT'21              &~~ 91.4 / 87.3	~~&~~ 63.0 / 60.7 ~~&~~	59.4 / 58.9	~~&~~ 96.7 / \underline{117.9} ~~&~~ 65.9 / \underline{76.0} ~~&~~ 57.4 / \underline{40.6} ~~ \\
                Liu \etal \cite{ref12_liu2020attention} CVPR'20         &~~ 84.2 / 78.8	~~&~~ 63.0 / 58.8 ~~&~~	64.8 / 67.9	~~&~~ 93.3 / 128.0 ~~&~~ 68.0 / 86.9 ~~&~~ \underline{58.8} / 40.2 ~~ \\
                \midrule
                Ours                                                    &~~ \textbf{64.1} / \textbf{59.8} ~~&~~ \textbf{48.0} / \textbf{44.7} ~~&~~ \textbf{77.3} / \textbf{79.5}	~~&~~ \textbf{69.1} / \textbf{63.1} ~~&~~ \textbf{49.7} / \textbf{45.9} ~~&~~ \textbf{74.7} / \textbf{77.8} ~~ \\
                \bottomrule
                \end{tabular}
            }
        	\caption{Comparison of performance on (distortion $d_1$) / (distortion $d_2$).}
    	\end{subtable}
        \begin{subtable}{\linewidth}
            \centering
            \small
            \tabcolsep=1mm
            \resizebox{0.83\textwidth}{!}{
                \begin{tabular}{@{}l|ccc|ccc@{}}
                \toprule
                & & \textit{Scenario 1} & & & \textit{Scenario 2} & \\
                \cmidrule{2-7}
                Method & MPJPE($\downarrow$) & P-MPJPE($\downarrow$) & PCKh@0.5($\uparrow$) & MPJPE($\downarrow$) & P-MPJPE($\downarrow$) & PCKh@0.5($\uparrow$)\\
                \midrule
                Martinez \etal \cite{ref7_martinez2017simple} ICCV'17   &~~ \underline{63.8} / 62.3 ~~&~~ 49.1 / 48.2 ~~&~~ 75.9 / 77.0 ~~&~~ 75.0 / 61.6 ~~&~~ 51.7 / 46.5 ~~&~~ 69.2 / 79.0 ~~ \\
                Zhao \etal \cite{ref8_zhao2019semantic} CVPR'19         &~~ 66.6 / 61.4 ~~&~~ 48.7 / 46.1 ~~&~~ 74.9 / 78.9 ~~&~~ 80.4 / 62.3 ~~&~~ 56.5 / 47.2 ~~&~~ 65.9 / 78.5 ~~ \\
                \midrule
                Pavllo \etal \cite{ref9_pavllo20193d} CVPR'19           &~~ 65.2 / 64.7 ~~&~~ 48.6 / 48.0 ~~&~~ \underline{76.5} / 76.9 ~~&~~ 74.8 / 54.2 ~~&~~ \underline{50.7} / 40.6 ~~&~~ 69.4 / 83.8 ~~ \\
                Chen \etal \cite{ref11_anatomy3D} TCSVT'21              &~~ 64.6 / \underline{60.7} ~~&~~ \underline{47.9} / \underline{44.7} ~~&~~ 76.3 / \underline{79.3} ~~&~~ 77.1 / \underline{53.1} ~~&~~ 52.9 / \underline{39.6} ~~&~~ 69.3 / \underline{85.2} ~~ \\
                Liu \etal \cite{ref12_liu2020attention} CVPR'20         &~~ 69.2 / 68.3 ~~&~~ 51.3 / 50.6 ~~&~~ 74.4 / 74.9 ~~&~~ \underline{72.3} / 55.6 ~~&~~ \underline{50.7} / 42.3 ~~&~~ \underline{70.4} / 83.2 ~~ \\
                \midrule
                Ours                                                    &~~ \textbf{53.8} / \textbf{53.4} ~~&~~ \textbf{40.8} / \textbf{40.4} ~~&~~ \textbf{83.1} / \textbf{83.4} ~~&~~ \textbf{51.8} / \textbf{51.4} ~~&~~ \textbf{39.5} / \textbf{38.9} ~~&~~ \textbf{85.6} / \textbf{85.8} ~~ \\
                \bottomrule
                \end{tabular}
            }
        	\caption{Comparison of performance on (distortion $d_3$) / (distortion $d_4$).}
    	\end{subtable}
    	\caption{Comparison with other state-of-the-art models on Human3.6M. The top two rows~\cite{ref7_martinez2017simple,ref8_zhao2019semantic} are based on a single-frame and others~\cite{ref9_pavllo20193d,ref11_anatomy3D,ref12_liu2020attention}, including our method, are based on video with a frame length of 27. Best in bold, second-best underlined.}
        \label{tbl:supp_quantitative_all_distortion}
    \end{table*}

    \begin{table*}[ht]
        \begin{subtable}{\linewidth}
            \centering
    		\small
        	\tabcolsep=1mm
        	\resizebox{\textwidth}{!}{
        		\begin{tabular}{@{}l|rrrrrrrrrrrrrrr|r@{}}
        		& Dir. & Disc. & Eat & Greet & Phone & Photo & Pose & Purch. & Sit & SitD. & Smoke & Wait & WalkD. & Walk & WalkT. & Avg\\
        		\midrule
        		Martinez \etal \cite{ref7_martinez2017simple} ICCV'17   & 80.5 & 79.9 & 70.2 & 74.3 & 69.7 & 58.3 & 74.7 & 79.5 & 67.6 & 56.4 & \underline{72.2} & \underline{73.5} & 65.4 & 78.8 & 72.5 & 71.6 \\
        		Zhao \etal \cite{ref8_zhao2019semantic} CVPR'19         & 75.3 & 73.1 & 69.8 & 73.1 & 70.9 & 57.4 & 70.6 & 76.5 & \underline{72.1} & \underline{59.7} & 71.6 & 70.7 & 65.8 & 73.9 & 70.3 & 70.1 \\
        		\midrule
        		Pavllo \etal \cite{ref9_pavllo20193d} CVPR'19           & \underline{82.0} & \underline{80.1} & \underline{76.0} & \underline{76.1} & \underline{71.7} & \underline{60.5} & \underline{75.5} & \underline{81.1} & 58.0 & 44.0 & 71.3 & 71.4 & 70.4 & \underline{82.7} & 78.8 & \underline{72.0} \\
        		Chen \etal \cite{ref11_anatomy3D} TCSVT'21              & 73.9 & 74.9 & 66.9 & 70.7 & 67.0 & 59.9 & 70.5 & 72.6 & 66.7 & 56.2 & 69.8 & 70.0 & 64.1 & 73.6 & 70.1 & 68.5 \\
        		Liu \etal \cite{ref12_liu2020attention} CVPR'20         & 80.6 & 78.3 & 72.5 & 74.1 & 70.5 & 59.7 & 74.9 & 79.9 & 52.9 & 41.0 & 69.7 & 71.3 & \underline{70.5} & 82.6 & \underline{79.0} & 70.5 \\
        		\midrule
        		Ours & \textbf{85.8} & \textbf{83.5} & \textbf{80.1} & \textbf{84.8} & \textbf{81.6} & \textbf{70.5} & \textbf{80.8} & \textbf{85.8} & \textbf{78.6} & \textbf{57.7} & \textbf{83.3} & \textbf{80.7} & \textbf{77.1} & \textbf{93.0} & \textbf{89.3} & \textbf{80.8} \\
        		\bottomrule
        		\end{tabular}
        	}
        	\caption{Reconstruction accuracy (PCKh@0.5) under the \textit{Scenario 1} setting.}
    	\end{subtable}
        \begin{subtable}{\linewidth}
            \centering
    		\small
        	\tabcolsep=1mm
        	\resizebox{\textwidth}{!}{
        		\begin{tabular}{@{}l|rrrrrrrrrrrrrrr|r@{}}
        		& Dir. & Disc. & Eat & Greet & Phone & Photo & Pose & Purch. & Sit & SitD. & Smoke & Wait & WalkD. & Walk & WalkT. & Avg\\
        		\midrule
        		Martinez \etal \cite{ref7_martinez2017simple} ICCV'17   & 59.2 & 61.0 & 62.0 & 57.1 & 62.2 & 48.9 & 58.3 & 67.6 & 69.6 & 65.5 & 63.1 & 57.9 & 60.4 & 59.9 & 57.2 & 60.7 \\
        		Zhao \etal \cite{ref8_zhao2019semantic} CVPR'19         & 57.3 & 57.7 & 59.2 & 55.4 & 61.2 & 47.1 & 56.7 & 63.2 & 65.2 & 62.9 & 61.6 & 56.0 & 59.0 & 59.4 & 56.5 & 58.6 \\
        		\midrule
        		Pavllo \etal \cite{ref9_pavllo20193d} CVPR'19           & 58.0 & 57.6 & 63.9 & 58.1 & 66.0 & 52.7 & 56.8 & 70.1 & 72.4 & 67.8 & 65.9 & 57.7 & 63.7 & 62.9 & 60.0 & 62.2 \\
        		Chen \etal \cite{ref11_anatomy3D} TCSVT'21              & 57.0 & 57.9 & \underline{66.1} & \underline{59.6} & \underline{66.7} & \underline{55.4} & 56.3 & 70.5 & \underline{72.7} & \underline{70.0} & \underline{66.8} & 58.8 & 63.6 & 64.3 & 61.3 & \underline{63.1} \\
        		Liu \etal \cite{ref12_liu2020attention} CVPR'20         & \underline{62.2} & \underline{62.2} & 63.7 & 58.9 & 65.6 & 51.7 & \underline{58.9} & \underline{70.7} & 70.1 & 66.8 & 65.9 & \underline{59.9} & \underline{64.0} & \underline{64.6} & \underline{61.9} & \underline{63.1} \\
        		\midrule
        		Ours & \textbf{82.0} & \textbf{82.0} & \textbf{74.4} & \textbf{76.7} & \textbf{82.1} & \textbf{63.6} & \textbf{80.1} & \textbf{82.4} & \textbf{80.5} & \textbf{66.7} & \textbf{80.8} & \textbf{77.3} & \textbf{75.0} & \textbf{86.0} & \textbf{81.7} & \textbf{78.1} \\
        		\bottomrule
        		\end{tabular}
        	}
        	\caption{Reconstruction accuracy (PCKh@0.5) under the \textit{Scenario 2} setting.}
    	\end{subtable}
    	\caption{Comparison with other state-of-the-art models on Human3.6M. The top two rows~\cite{ref7_martinez2017simple,ref8_zhao2019semantic} are based on a single-frame and others~\cite{ref9_pavllo20193d,ref11_anatomy3D,ref12_liu2020attention}, including our method, are based on video with a frame length of 27. The reported performance is the average value for all kinds of distortions. Higher is better, best in bold, second-best underlined.}
        \label{tbl:supp_quantitative_actionwise}
    \end{table*}

    \begin{figure}[t]
    	\begin{center}
    		\captionsetup{justification=centering}
    		\begin{subfigure}[t]{\linewidth}
    			\centering
    			\includegraphics[width=1\columnwidth]{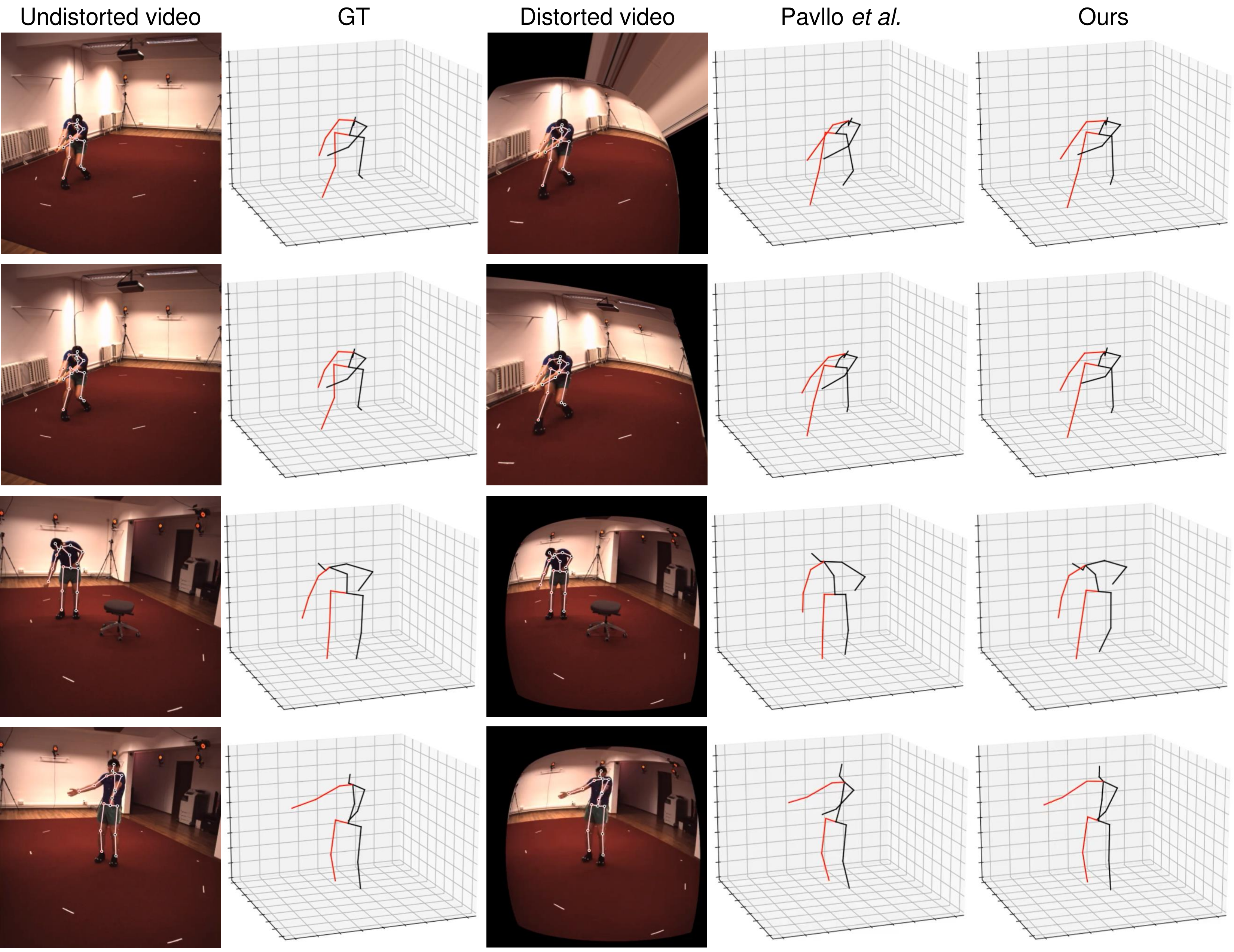}
    			\caption{Qualitative results under the \small\textit{Scenario 1} \normalsize setting.}
    		\end{subfigure}
    		\begin{subfigure}[t]{\linewidth}
    			\centering
    			\includegraphics[width=1\columnwidth]{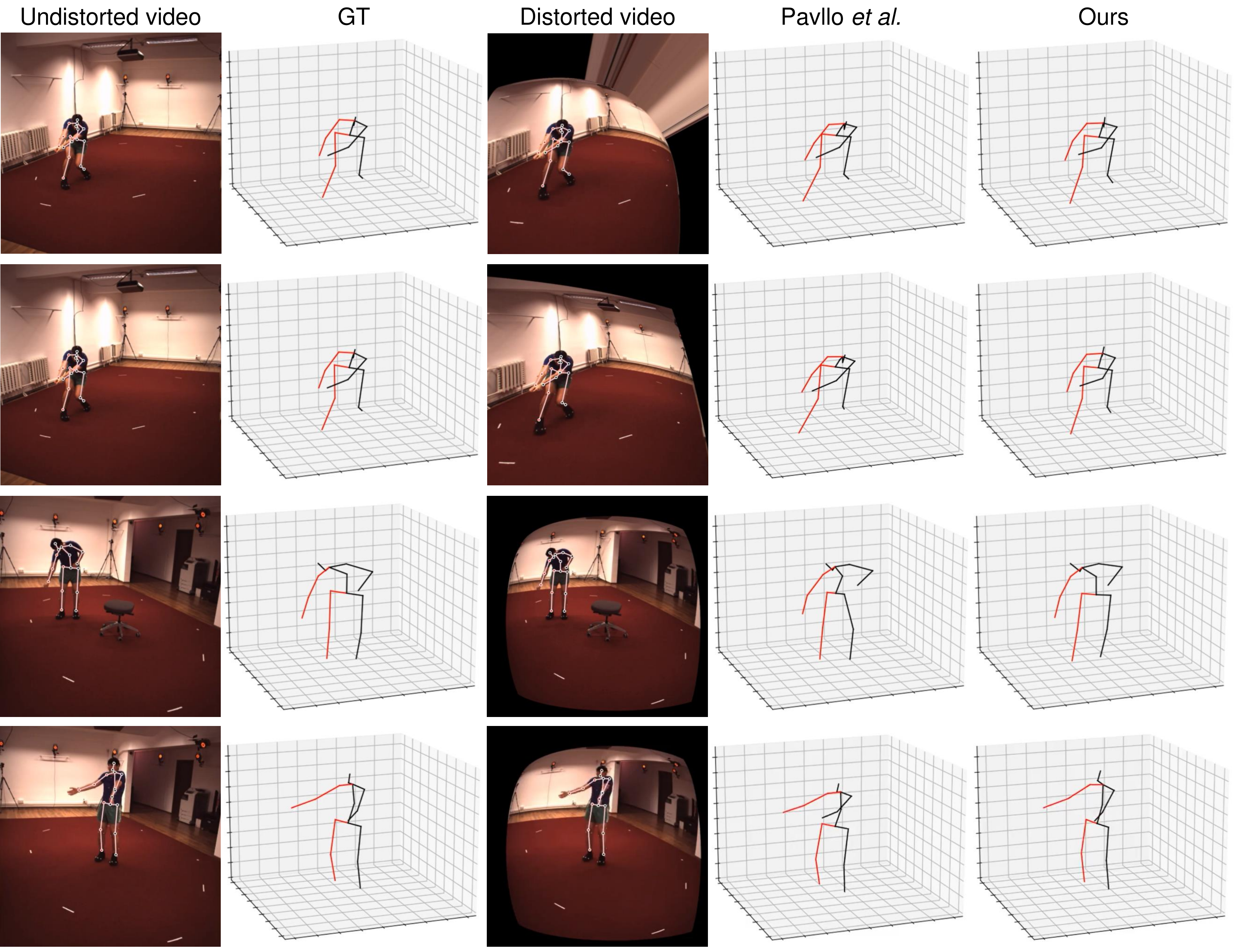}
    			\caption{Qualitative results under the \small\textit{Scenario 2} \normalsize setting.}
    		\end{subfigure}
    	\end{center}
    	    \vspace{-4mm}
    		\caption{Qualitative results on heavily distorted videos of Human3.6M under the \small\textit{Scenario 1} \normalsize and \small\textit{Scenario 2} \normalsize setting.}
    	\label{fig:supp_qualitative}
    	\vspace{-1mm}
    \end{figure}

    \subsection{Qualitative Results}
        \label{supp:qualitative results}
        Figure~\ref{fig:supp_qualitative} shows qualitative results from videos with more diverse poses and distortions. We can notice that our method adapts better to the distorted environments than our base model~\cite{ref9_pavllo20193d}, showing more similar results to the ground-truth 3D pose.

    \subsection{Performance in Undistorted Environments}
        Since our model is trained to be sensitive to all kinds of distortions, it performs well even in undistorted environments. Our method shows an MPJPE of 50.6mm in the test environment with no distortion. This is 2.1mm higher than the base model \cite{ref9_pavllo20193d}, but it is a reasonable trade-off because it has great advantages in other situations with distortions.

\makeatletter
\setlength{\@fptop}{0pt}
\makeatother